\documentclass[journal]{IEEEtran}

\usepackage{url}

\usepackage{color}
\usepackage{amsmath}
\usepackage{todonotes}
\usepackage{xspace}
\usepackage{subfigure}
\usepackage{cite}
\usepackage{enumitem}

\newtheorem{problem}{Problem}
\newcommand{\MM}{\textsc{M-M}\xspace}
\newcommand{\MDTU}{\textsc{M-DTU}\xspace}
\newcommand{\MF}{\textsc{M-F}\xspace}
\newcommand{\FM}{\textsc{F-M}\xspace}
\newcommand{\FDTU}{\textsc{F-DTU}\xspace}
\newcommand{\FF}{\textsc{F-F}\xspace}
\newcommand{\MMDU}{\textsc{M-MDU}\xspace}
\newcommand{\MDUM}{\textsc{M-DUM}\xspace}
\newcommand{\MDUMDU}{\textsc{M-DUMDU}\xspace}
\newcommand{\FFDU}{\textsc{F-FDU}\xspace}
\newcommand{\FDUF}{\textsc{F-DUF}\xspace}
\newcommand{\FDUFDU}{\textsc{F-DUFDU}\xspace}
\newcommand{\MFDU}{\textsc{M-FDU}\xspace}
\newcommand{\MDUF}{\textsc{M-DUF}\xspace}
\newcommand{\MDUFDU}{\textsc{M-DUFDU}\xspace}
\newcommand{\FMDU}{\textsc{F-MDU}\xspace}
\newcommand{\FDUM}{\textsc{F-DUM}\xspace}
\newcommand{\FDUMDU}{\textsc{F-DUMDU}\xspace}
\newcommand{\BC}{boustrophedon cell\xspace}
\newcommand{\BCS}{boustrophedon cells\xspace}
\newcommand{\CON}{Concorde\xspace}

\newcommand{\norm}[1]{\left\lVert#1\right\rVert_2}

\begin{document}

\title{\LARGE \bf Coverage of an Environment Using\\Energy-Constrained Unmanned Aerial Vehicles}

\author{Kevin Yu, \and Jason M. O'Kane, \and Pratap Tokekar
\thanks{This material is based upon work supported by the National Science Foundation under Grant No. 1526862, 1513203, 156624, 1637915, \& 1849291. Yu is  with the Department of Electrical \& Computer Engineering, Virginia Tech, U.S.A. Email: \texttt{\small klyu@vt.edu} \quad O'Kane is with the Department of Computer Science \& Engineering, University of South Carolina, U.S.A. \texttt{\small jokane@cse.sc.edu} \quad Tokekar is with the Department of Computer Science, University of Maryland, U.S.A. Email: \texttt{\small tokekar@umd.edu}}}

\maketitle
\IEEEpeerreviewmaketitle

\begin{abstract}
We study the problem of covering an environment using an Unmanned Aerial Vehicle (UAV) with limited battery capacity. We consider a scenario where the UAV can land on an Unmanned Ground Vehicle (UGV) and recharge the onboard battery. The UGV can also recharge the UAV while transporting the UAV to the next take-off site. We present an algorithm to solve a new variant of the area coverage problem that takes into account this symbiotic UAV and UGV system. The input consists of a set of \emph{boustrophedon cells} --- rectangular strips whose width is equal to the field-of-view of the sensor on the UAV. The goal is to find a coordinated strategy for the UAV and UGV that visits and covers all cells in minimum time, while optimally finding how much to recharge, where to recharge, and when to recharge the battery. This includes flight time for visiting and covering all cells, recharging time, as well as the take-off and landing times. We show how to reduce this problem to a known NP-hard problem, Generalized Traveling Salesperson Problem (GTSP). Given an optimal GTSP solver, our approach finds the optimal coverage paths for the UAV and UGV. Our formulation models multi-rotor UAVs as well as hybrid UAVs that can operate in fixed-wing and Vertical Take-off and Landing modes. We evaluate our algorithm through simulations and proof-of-concept experiments.
\end{abstract}

\renewcommand{\abstractname}{Note to Practitioners}
\begin{abstract}
There are many applications, such as environmental monitoring, where Unmanned Aerial Vehicles (UAVs) can automate data collection by covering the environment using a mobile sensor. If the environment is large, then it may be infeasible to cover it with an energy-constrained UAV. An option would be to replace the batteries of the UAV along the flight or use a recharging station to aid complete coverage. These alternatives can be limited since they may require manual intervention or inefficient flights back-and-forth between the charging stations. Instead, we present a new approach that uses an Unmanned Ground Vehicle (UGV) as a mobile recharging station. We allow for the UAV wants to autonomously rendezvous with the UGV, land on it, recharge, and potentially be transported to another location before taking-off. The environment to be monitored is given as input in the form of a set of rectangular strips that need to be covered in minimum time with one or more recharging stops. We present an algorithm to find the optimal solution to this problem and verify the performance through simulations.
\end{abstract}

\section{Introduction}
There are many applications such as infrastructure inspection and environmental monitoring~\cite{liu2014review,ozaslaninspection,dunbabin2012environmental}, surveillance~\cite{michael2011persistent}, precision agriculture~\cite{das2015devices,tokekar2016sensor}, and search and rescue~\cite{sherman2018cooperative,sung2017algorithm} where Unmanned Aerial Vehicles (UAVs) can be used as mobile, adaptive sensors. A specific use-case in such applications is to provide visual aerial coverage with UAVs. The key challenge we address in this paper is how to cover large environments with an energy-constrained UAV. Particularly, we are interested in scenarios where the coverage is expected to take longer than the battery runtime of the UAV. 

One way of addressing energy constraints is by choosing a better platform. Multi-rotor UAVs have limited battery runtime, typically less than 30 minutes~\cite{10BestLo87:online}. As a result, surveying large areas with a single vehicle may require frequent stops to recharge or replace batteries. Fixed-wing UAVs have longer runtimes, typically less than 90 minutes, but cannot take-off and land vertically or hover in place. The latter characteristic may be essential for visual coverage. Fixed-wing UAVs also have the steering constraints that limit their maneuverability. Hybrid UAVs seek to achieve the best of both worlds --- higher maneuverability of a multi-rotor and longer endurance of a fixed-wing. Such hybrid UAVs are commercially available for coverage applications such as precision agriculture~\cite{AVDroneA85:online}, environmental monitoring~\cite{WingtraO65:online} reconnaissance~\cite{Quantix67:online} that involve visual coverage of large areas.

While fixed-wing and hybrid UAVs can mitigate some of the energy limitations, there may still be environments that are too large or need persistent monitoring beyond the runtime of the UAV. To address this inherit limitation, we propose a solution that uses Unmanned Ground Vehicles (UGVs) as mobile recharging stations. In our prior work~\cite{yu2018autonomous}, we presented an approach to visit a set of specified points of interest using a multi-rotor UAV with a UGV acting as a mobile recharging station. In this paper, we investigate the coverage problem using hybrid UAVs.

\begin{figure}[t]
\centering
\includegraphics[width = 0.9\columnwidth]{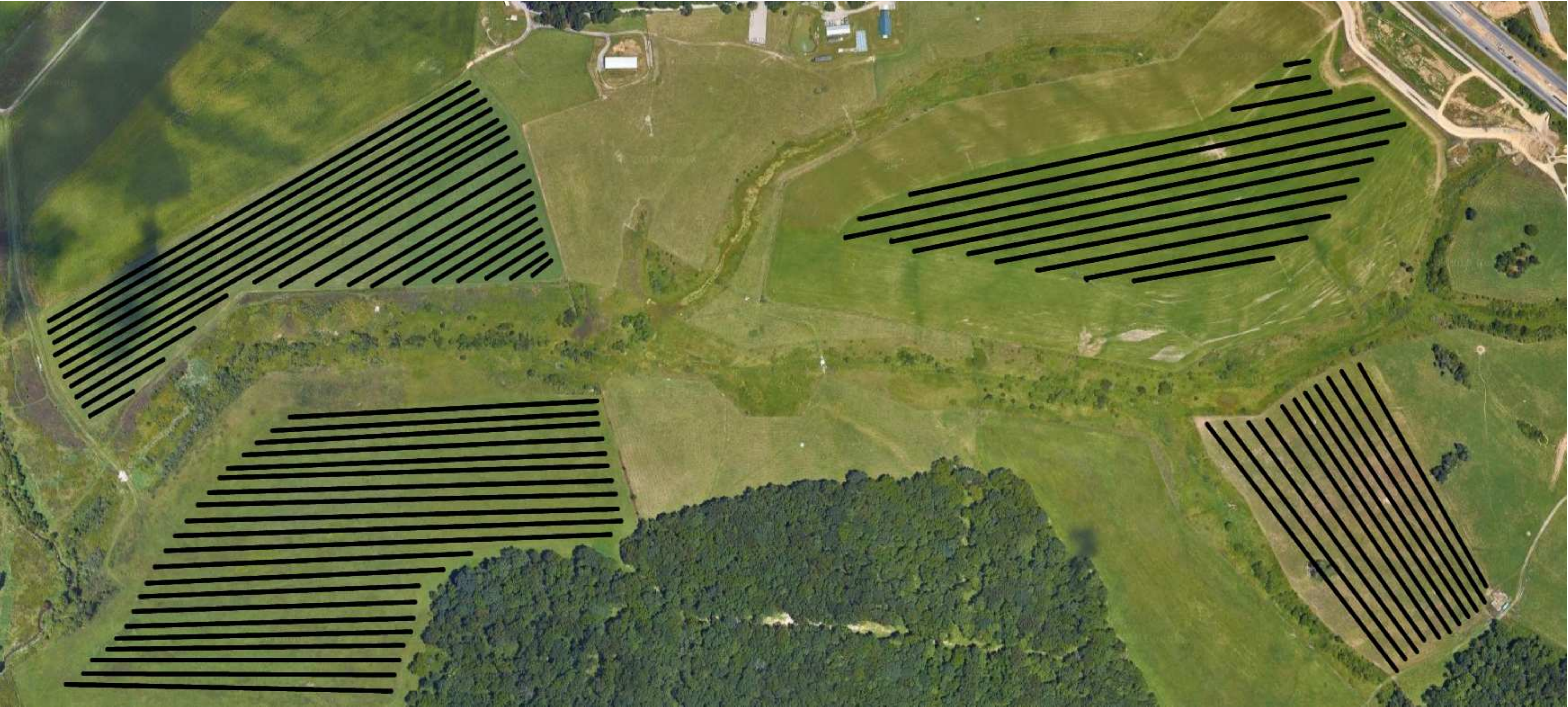}
\caption{We study the problem of covering a set of boustrophedon cells using a UAV which has limited battery capacity, but multiple modes of flight. In a precision agriculture scenario, a boustrophedon cell may correspond to a row of crops that must be monitored using a downward-facing camera on the UAV.\label{fig:BCS}}
\end{figure}

The input to our planner is a set of \emph{boustrophedon cells} --- rectangular regions whose width is equal to the footprint of the UAV's sensor. The boustrophedon cells can be obtained by decomposing regions that need to be covered~\cite{choset1997coverage,maza2007multiple}. Figure~\ref{fig:BCS} shows a motivating example of surveying crops in four fields. Here, each \BCS corresponds to a row of crops that need to be imaged by the UAV. 

A \BC can be covered by the UAV entering from either end and exiting from the other --- the planner must find the optimal sequence in which to cover the \BCS as well as the corresponding entry and exit sites for each \BCS.

We consider scenarios where the UAV can land on the UGV and either recharge in-place or recharge while the UGV transports the UAV to the next take-off site. We present an algorithm that plans a tour for the UAV and a path for the UGV, such that the UAV can cover an area in the minimum time while never running out of charge. This includes not only the flight time of the UAV but also the time it takes to recharge as well as the taking-off and landing times. The output tour given by our planner specifies not only the order in which to cover the \BCS but also the charging schedule and flight mode that the UAV will be in. In particular, the planner outputs where and how to recharge the battery, how much to recharge by, and whether to fly in multi-rotor flight mode or fixed-wing flight mode.



This problem is a generalization of the NP-hard Traveling Salesperson Problem (TSP)~\cite{arora1998polynomial}. Our solution is based on reducing the coverage problem to the Generalized TSP (GTSP)~\cite{noon1993efficient}. Specifically, we present an algorithm that is guaranteed to find the optimal coverage tour for the UAV, as long the GTSP solver produces optimal tours. While no polynomial-time optimal algorithms are believed to exist for NP-hard problems, there are solvers that find optimal solutions to many instances in reasonable amounts of time in practice. We use one such solver, Generalized Large Neighborhood Search (GLNS)~\cite{Smith2016GLNS}, that finds optimal solutions for GTSP instances. We empirically evaluate the performance of our algorithm using the GLNS solver.


This work extends our prior work~\cite{yu2018autonomous} wherein we presented a method to visit a set of points, instead of regions, using a multi-rotor. This journal article extends the preliminary work presented in~\cite{yu2019coverage} wherein we presented a method to visit a set of \BCS using a multi-rotor UAV and UGV as a mobile recharging station. In this paper, we extend the results to account for a hybrid UAV. Hybrid UAVs add the challenges of handling multiple modes of flight, Dubins' steering constraints when executing fixed-wing flight, and the different energy discharge rates dependent on the flight mode.

\section{Related Work}\label{sec:related}
Environmental coverage is a well-studied problem in robotics~\cite{Choset2001, GalCar13}. The variant most closely related to the one we study is that of decomposing a known environment into various cells and then finding a route to sweep (i.e., cover) each cell~\cite{huang2001, yao2006, gonzalez2005bsa, choi2009online, bochkarev2016minimizing}. In this work, we assume that the first step (cell decomposition) has been solved and focus on the problem of routing the UAV to cover the cells in minimal time, while keeping track of the battery level and flight mode. We introduce a novel means of environmental coverage using a hybrid UAV system, which can leverage aspects of a multi-rotor UAV, such as vertical take-off and landing, and fixed-wing UAV, such as long flight times. Specifically, we take as input a boustrophedon cell decomposition which can be found using techniques given by~\cite{maza2007multiple}.

A number of algorithms have also been developed for the second step, i.e., routing to cover all cells, under various constraints. In particular, Karapetyan et al. presented a multi-robot coverage algorithm for boustrophedon cell decomposition for point robots~\cite{RekleitisIROS2017b}. Yu et al. presented a coverage algorithm for a single robot with Dubins steering constraints~\cite{YuRopHun15}. Lewis et al. later proved that problem NP-hard, and showed how to reduce the graph to obtain practical solutions~\cite{LewEdw17}. Bochkarev and Smith present a method for decomposing an environment to minimize the number of turns needed to cover an area, but do not consider energy-limited robots and consequently, do not need to keep track of the battery level of the robot~\cite{bochkarev2016minimizing}. Most recently, Karapetyan et al. presented two techniques to cover a collection of cells with multiple Dubins vehicles~\cite{KarMou18}.

The underlying ideas in the aforementioned works are similar --- reduce the problem of covering all cells to a variant of the TSP, solve the TSP, and convert the resulting solution back to a tour for the robots. Our approach extends these ideas for the case of a robot with limited energy capacity which can be recharged along the way and a robot with multiple flight modes, taking advantage of both multi-rotor and fixed-wing modalities. This requires us to keep track of energy level of the robot along the tour and the flight modes, which further complicates the problem. Nevertheless, we show that by reducing it to GTSP, we can obtain optimal solutions in reasonable amounts of time.

Many other works also analyze the coverage problem using fixed-wing UAVs, but do not have the advantages of a hybrid system. Paull et al. considered area coverage using onboard sensors for a fixed-wing UAV~\cite{paull2013sensor}. However the authors take an online approach, leading to solutions that are sub-optimal for environmental coverage. Xu et al. present an algorithm for optimal terrain coverage using fixed-wing UAV~\cite{xu2011optimal}. Their method considers minimizing overlapping areas of coverage, the method in this paper does not allow for any overlapping coverage. Also we provide experiments that utilize charging stations and implement a hybrid system. Coombes et al. use a fixed-wing UAV for survey coverage path planning in windy environments~\cite{coombes2018fixed}. This approach studies the effects of wind on a fixed-wing system and proposes algorithms that solve for paths that take into account the wind patterns of the environment.

There have been algorithms for assigning and routing with one or more stationary recharging stations~\cite{kim2013scheduling, ahmed2016energy,liu2014energy}. Kim et al. present a Mixed Integer Linear Programming approach for assigning UAVs to stationary recharging stations while taking into account the task objective~\cite{kim2013scheduling}. Liu and Michael presented a method for assigning UAVs to UGVs acting as recharging stations~\cite{liu2014energy}.

In our previous work~\cite{tokekar2016sensor,yu2018autonomous,yu2019coverage}, we studied the problem of visiting a set of boustrophedon cells (rectangles with the width of the field-of-view (FOV) of the sensor in a 2D plane) using an energy-limited UAV with only one mode of flight. In~\cite{tokekar2016sensor}, we showed how to maximize the number of sites visited in a single charge when the UAV is allowed to land on the UGV and let the UGV transport it to the next take-off site without the UAV expending energy. In~\cite{yu2018autonomous}, we extended this to also allow for the UGV to recharge the UAV either while stationary or while being transported to the next deployment site. In~\cite{yu2019coverage}, we further extend the work from~\cite{yu2018autonomous} to conduct coverage of boustrophedon cells. This paper extends the prior work from merely having a single mode of flight to having multiple modes of flight. As a result, the planner must decide not only the order in which the boustrophedon cells should be visited but also the directions in which to cover them and what is the optimal flight mode to use.

\section{Problem Formulation}\label{sec:prob}
\begin{figure}[ht]
    \centering
    \includegraphics[width = \columnwidth]{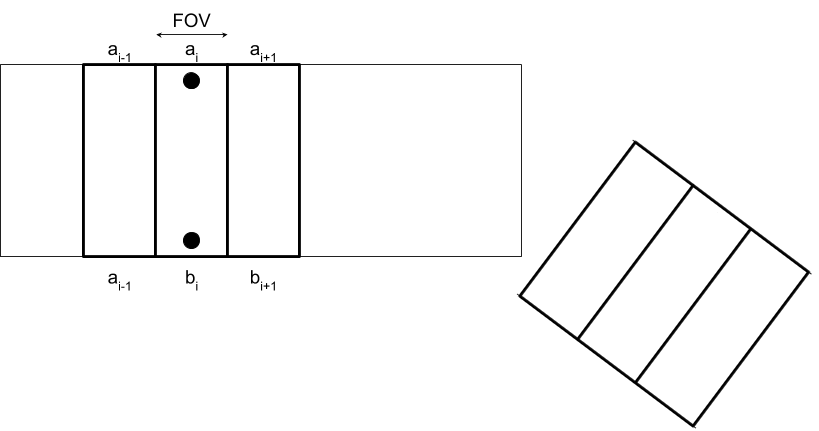}
    \caption{Example \BCS. Each \BC is a rectangle whose width is equal to the footprint of the UAV's sensor. A \BC $i$ is characterized by two sites, $a_i$ and $b_i$, on either end. The algorithm must choose which one acts as the entry site and how to traverse the boustrophedon cell.\label{fig:BCE}}
\end{figure}

The input to our algorithm is a set of $n$ boustrophedon cells that need to be covered by the UAV. A \BC is a rectangular strip whose width is equal to the diameter of the FOV of the sensor onboard the robot. An example is shown in Figure~\ref{fig:BCE} and a larger example is shown in Figure~\ref{fig:BCS}.

Each \BC is described by two sites, $a_i$ and $b_i$, where $i$ is the index of the \BC. These sites are placed at two ends of the rectangular strip. A \BC is said to be \emph{covered} if the UAV travels in a straight line from $a_i$ to $b_i$ or from $b_i$ to $a_i$. The UAV can enter a \BC from either site, $a_i$ or $b_i$. However, once the UAV has entered a \BC it is required to cover the entire \BC and exit from the other site. The coverage algorithm must choose one of the sites as the entry site as well as how to traverse the \BC. We slightly generalize the traditional notion of a \BC by allowing them to be oriented in different directions. That is, we do not require the \BCS to be parallel to each other.

We make the following assumptions:
\begin{itemize}[leftmargin=*]
\item the UAV has an initial battery charge of $100\%$;
\item the UAV flies at a fixed-altitude plane when covering a \BC;
\item the UAV travels at unit speed in multi-rotor mode;
\item the UGV has unlimited fuel/battery capacity;
\item edges can only conduct recharging if they correspond with sites the UGV can visit.
\end{itemize}

We use $t_{TO}$ and $t_L$ to represent the time taken by the UAV to take-off from the UGV to reach the fixed-altitude plane and land from this plane onto the UGV, respectively. $D_{\max}$ represents the total distance a UAV can travel with 100\% battery capacity in multi-rotor mode.\footnote{Strictly speaking, we maintain a reserve battery capacity to take-off from ground to reach the fixed altitude plane and land from the fixed-altitude plane on the ground. In this paper, when we refer to 100\% battery capacity, it excludes this reserve battery for taking-off and landing.} We discretize the battery capacity into $C$ levels. $r$ represents the rate the battery gets recharged per unit time. $fRatio$ represents the ratio of multi-rotor to fixed-wing battery consumption. This is above $1$ if the fixed-wing consumes less battery than the multi-rotor over a fixed distance. We define the turn radius of the fixed-wing as $TR$. When our UAV is in fixed-wing mode it will obey the constraints of a Dubins vehicle. 


Let $\gamma(i) \in \{a_i,b_i\}$ denote the site chosen by a coverage algorithm to be the entry site of the $i^{th}$ \BC in the order in which they are to be visited. Correspondingly, $\overline{\gamma}(i)$ denotes the site chosen to be the exit site of the \BC, i.e., $\overline{\gamma}(i) = \{a_i,b_i\}\backslash\gamma(i)$. 
$\sigma(j)$ denotes the order in which the \BCS are to be visited. That is, $\sigma(j) \in \{1, \ldots, n\}$ gives the $j^{th}$ \BC that is visited. 

We use $\gamma_i$, $\overline{\gamma}_{i}$, and $\gamma_{i+1}$ to denote $\gamma(\sigma(i))$, $\overline{\gamma}(\sigma(i))$, and $\gamma(\sigma(i+1))$, respectively. We denote by $t_{G}(\overline{\gamma}_j, \gamma_{j+1})$ the time it takes for the UGV to travel along the ground from the exit site of $j^{th}$ \BC to the entry site of the next visited \BC. We use $t_{M}(\overline{\gamma}_j, \gamma_{j+1})$ to represent the time it takes for the UAV to travel in multi-rotor mode from the exit site of the $j^{th}$ \BC to the entry site of the next \BC visited. Similarly, $t_{M}(\gamma_j, \overline{\gamma}_j)$ gives the time taken by the UAV in multi-rotor mode to cover the $j^{th}$ \BC. We use $t_{F}(\overline{\gamma}_j, \gamma_{j+1})$ to represent the time it takes for the UAV to travel in fixed-wing mode from the exit site of the $j^{th}$ \BC to the entry site of the next \BC visited. Similarly, $t_{F}(\gamma_j, \overline{\gamma}_j)$ gives the time taken by the UAV in fixed-wing mode to cover the $j^{th}$ \BC.

Suppose $\pi$ is a path that visits every \BC in the order given by $\sigma$ and with entry and exit sites given by $\gamma$. The cost of the path depends on how the UAV travels between consecutive \BCS. Consider traveling from $\gamma_j$ to $\overline{\gamma}_j$ and then on to $\gamma_{j+1}$ along $\pi$. We have the following components for this part of the path:
\begin{itemize}[leftmargin=*]
    \item The UAV must fly from $\gamma_j$ to $\overline{\gamma}_j$. The time taken is given by $t_{M}(\gamma_j, \overline{\gamma}_j)$ or $t_{F}(\gamma_j, \overline{\gamma}_j)$. Let $I_1(\gamma_j,\overline{\gamma}_j)$ be an indicator function that denotes whether the UAV chooses to fly in multi-rotor or fixed-wing mode.
    \item It can then choose to land on the UGV at $\overline{\gamma}_j$, recharge in-place, and take-off to reach the fixed-altitude plane at $\overline{\gamma}_j$. Let $I_2(\overline{\gamma}_j)$ be an indicator function that denotes whether the UAV chooses to do this or not.
    \item It can then choose to either fly from $\overline{\gamma}_j$ to $\gamma_{j+1}$ or land on the UGV at $\overline{\gamma}_j$, recharge while being carried by the UGV to the next site, then take-off at $\gamma_{j+1}$ to reach the fixed-altitude plane. Let $I_3(\overline{\gamma}_j)$ be an indicator function denoting whether the UAV travels with the UGV or not. Note that if UAV chooses flight then the indicator function $I_1(\overline{\gamma}_j,\gamma_{j+1})$ is also used to denote true for multi-rotor or false for fixed-wing flight.
    \item It can then choose to land on the UGV at $\gamma_{j+1}$, recharge in-place, and take-off to reach the fixed-altitude plane at $\gamma_{j+1}$. Let $I_2(\gamma_{j+1})$ be an indicator function that denotes whether the UAV chooses to do this or not.
\end{itemize}
Based on these choices, the cost of traveling from $\gamma_j$ to $\gamma_{j+1}$ is given by:
\begin{equation}
\begin{split}
& T(j, j+1) = I_1(\gamma_j,\overline{\gamma}_j)t_{M}(\gamma_j, \overline{\gamma}_j)\\
& +(1-I_1(\gamma_j,\overline{\gamma}_j))t_{F}(\gamma_j, \overline{\gamma}_j)\\
& +I_2(\overline{\gamma}_j)(t_{L} + r\cdot b(\overline{\gamma}_j, \overline{\gamma}_j) + t_{TO})\\
& +(1-I_3(\overline{\gamma}_j))I_1(\overline{\gamma}_j,\gamma_{j+1})t_{M}(\overline{\gamma}_j, \gamma_{j+1})\\
& +(1-I_3(\overline{\gamma}_j))(1-I_1(\overline{\gamma}_j,\gamma_{j+1}))t_{F}(\overline{\gamma}_j, \gamma_{j+1})\\
& +I_3(\overline{\gamma}_j)(\max\{t_{G}(\overline{\gamma}_j, \gamma_{j+1}), r\cdot b(\overline{\gamma}_j, \gamma_{j+1})\}\\
& + t_{L} + t_{TO})+I_2(\gamma_{j+1})(t_{L} + r\cdot b(\gamma_{j+1},\gamma_{j+1}) + t_{TO}).\\
\end{split}
\label{eq:costPerOne}
\end{equation}
Here, $b(\cdot)$ is a function which gives the amount by which the battery should be recharged between two sites.

Therefore, we can define the cost of the path $\pi$ as:
\begin{equation}
T(\pi) = \sum_{j=1}^{n-1} T(\gamma_j,\gamma_{j+1}) + \min\{t_{M}(\gamma_n,\overline{\gamma}_n), t_{F}(\gamma_n,\overline{\gamma}_n)\}
\label{eq:costFun}
\end{equation}
At the end of $\pi$ we take the minimum of $t_M$ and $t_F$ because at the last site the UAV will not need to conduct any type of charging and therefore will only need to cover the site. We are now ready to define the problem.
\begin{problem}[Multiple Polygon Coverage]\label{pr:MPC}
Given a set of \BCS to be covered, find a path $\pi^*$, which contains $\sigma(\cdot)$, $\gamma(\cdot)$, $I_2(\cdot)$, $I_3(\cdot)$, $I_1(\cdot)$ and $b(\cdot)$, for the UAV that visits and covers all of the \BCS, while minimizing the cost (Equation~\ref{eq:costFun}), and ensuring that the UAV does not run out of battery capacity. The path $\pi^*$ must specify the order in which to visit the \BCS, $\sigma(\cdot)$, the entry site for each \BC, $\gamma(\cdot)$, the recharging indicator functions, $I_2(\cdot)$, $I_3(\cdot)$ and $I_1(\cdot)$, the amount of recharging at a site, $b(\cdot)$, and when to change flight modes during coverage.
\end{problem}

Note that finding a path for the UAV necessitates finding a path for the UGV that supports the UAV recharging schedule.

\section{GTSP-based Algorithm}\label{sec:gtsp}
We solve the polygon coverage problem by reducing it to GTSP. In this section, we describe in detail the reduction to GTSP. The input to GTSP is a directed weighted graph where the vertices are partitioned into clusters. The objective is to find a minimum cost tour that visits exactly one vertex in each cluster. If all the clusters contain exactly one vertex, then GTSP trivially reduces to TSP. We show how to create the clusters, the edges between the clusters and then show how to convert the solution for GTSP into tours for the UAV and the UGV.

\subsection{Vertices and Clusters}\label{sec:vertsandclusters}
We discretize the battery's charge states into $C$ levels. We create $C$ vertices, one corresponding to each discretized battery level, for each site $a_i$ and $b_i$ corresponding to \BC $i$. The vertex is denoted by $v_{a_i}^{k}$ or $v_{b_i}^{k}$, where $k$ corresponds to the discretized battery level. Thus there are $2nC$ total vertices in the graph. We create one cluster per \BC. This cluster contains $2C$ vertices, $C$ of them corresponding to $a_i$ and $C$ of them corresponding to $b_i$.

\subsection{Edges}\label{sec:edges}
We create an edge between every pair of vertices that do not belong to the same cluster (i.e., do not belong to the same \BC). Recall that a vertex corresponds to the entry site for the corresponding \BC. Therefore, an edge between two vertices represents the UAV starting at the entry site of the first \BC and ending at the entry site of the next \BC. This includes two travel legs: coverage of the first \BC and then traveling from the exit site of the first \BC towards the second \BC. Recall that the UAV must always fly the first leg, either in multi-rotor or fixed-wing mode; however, the second leg can be a combination of recharging, flying, and/or recharging while traveling on the UGV.

\begin{figure}[t]
\centering
\includegraphics[width = \columnwidth]{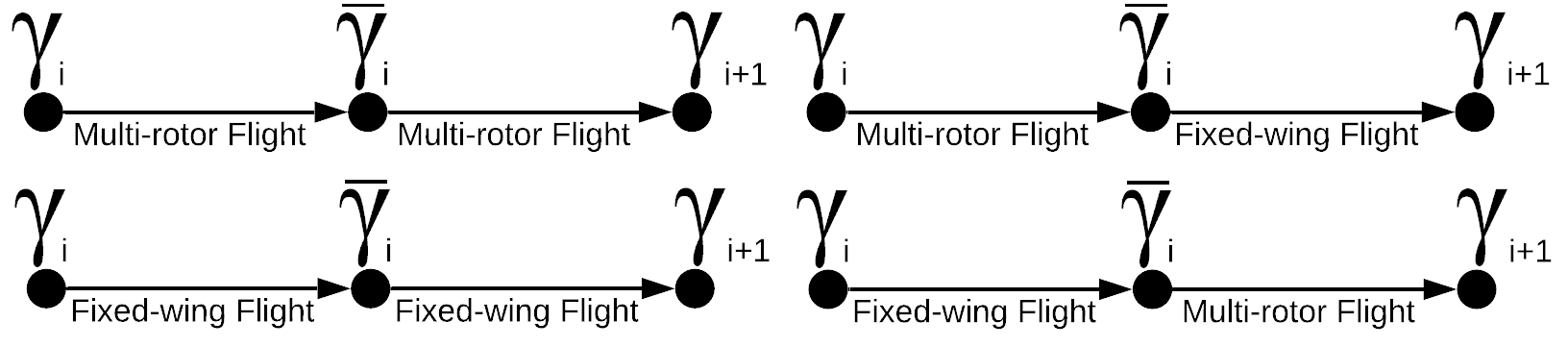}
\caption{\MM, \FF, \MF, \FM.}
\label{fig:xx}
\end{figure}

\begin{figure}[t]
\centering
\includegraphics[width = \columnwidth]{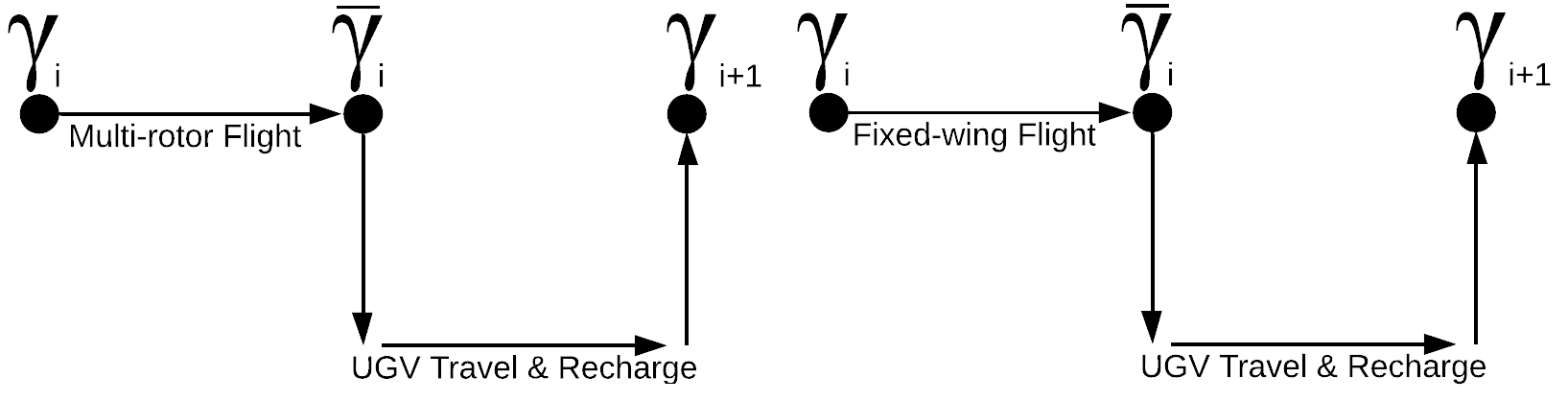}
\caption{\MDTU, \FDTU.}
\label{fig:xdtu}
\end{figure}

\begin{figure}[t]
\centering
\includegraphics[width = \columnwidth]{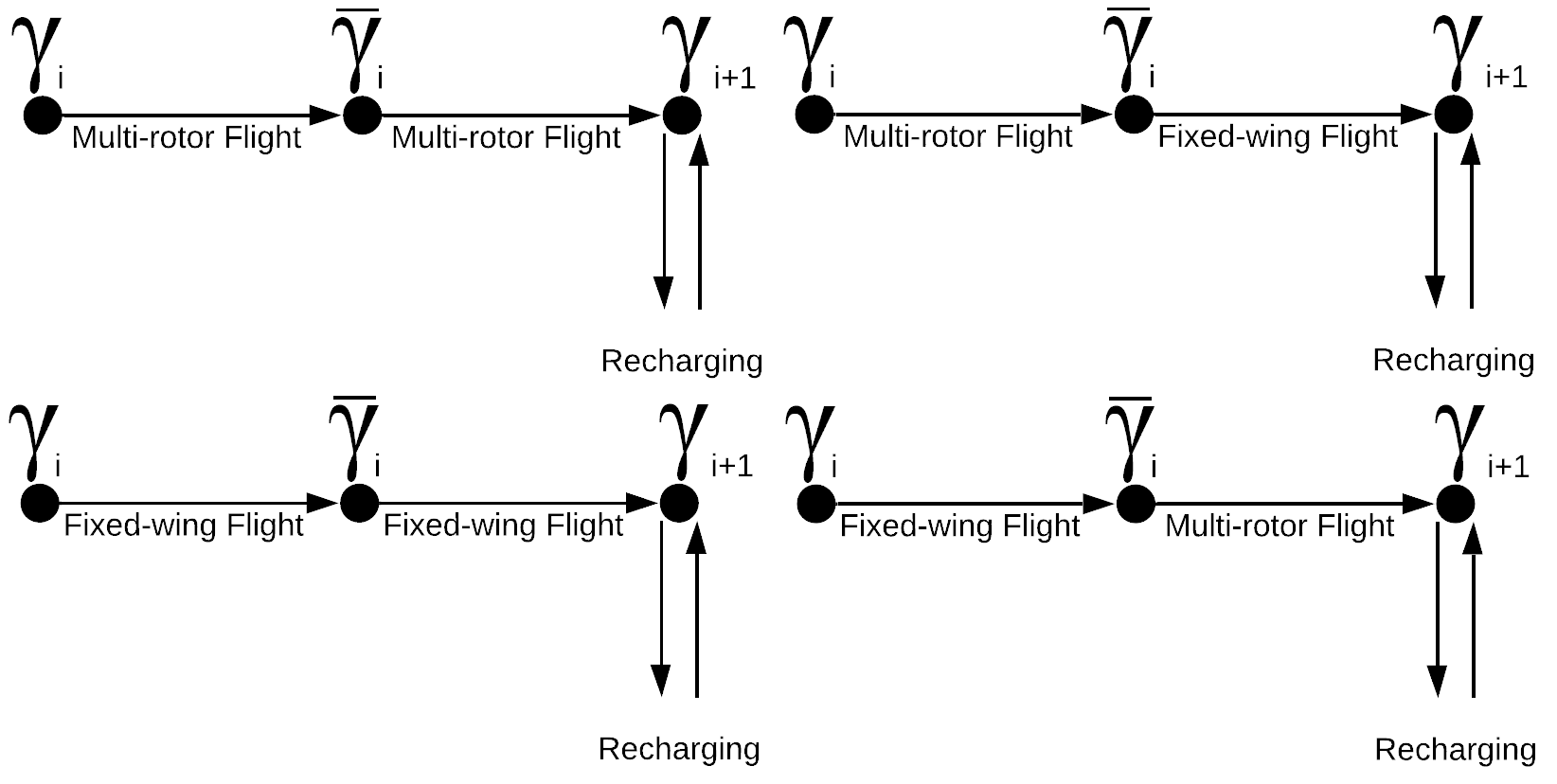}
\caption{\MMDU, \FFDU, \MFDU, \FMDU.}
\label{fig:xxdu}
\end{figure}

\begin{figure}[t]
\centering
\includegraphics[width = \columnwidth]{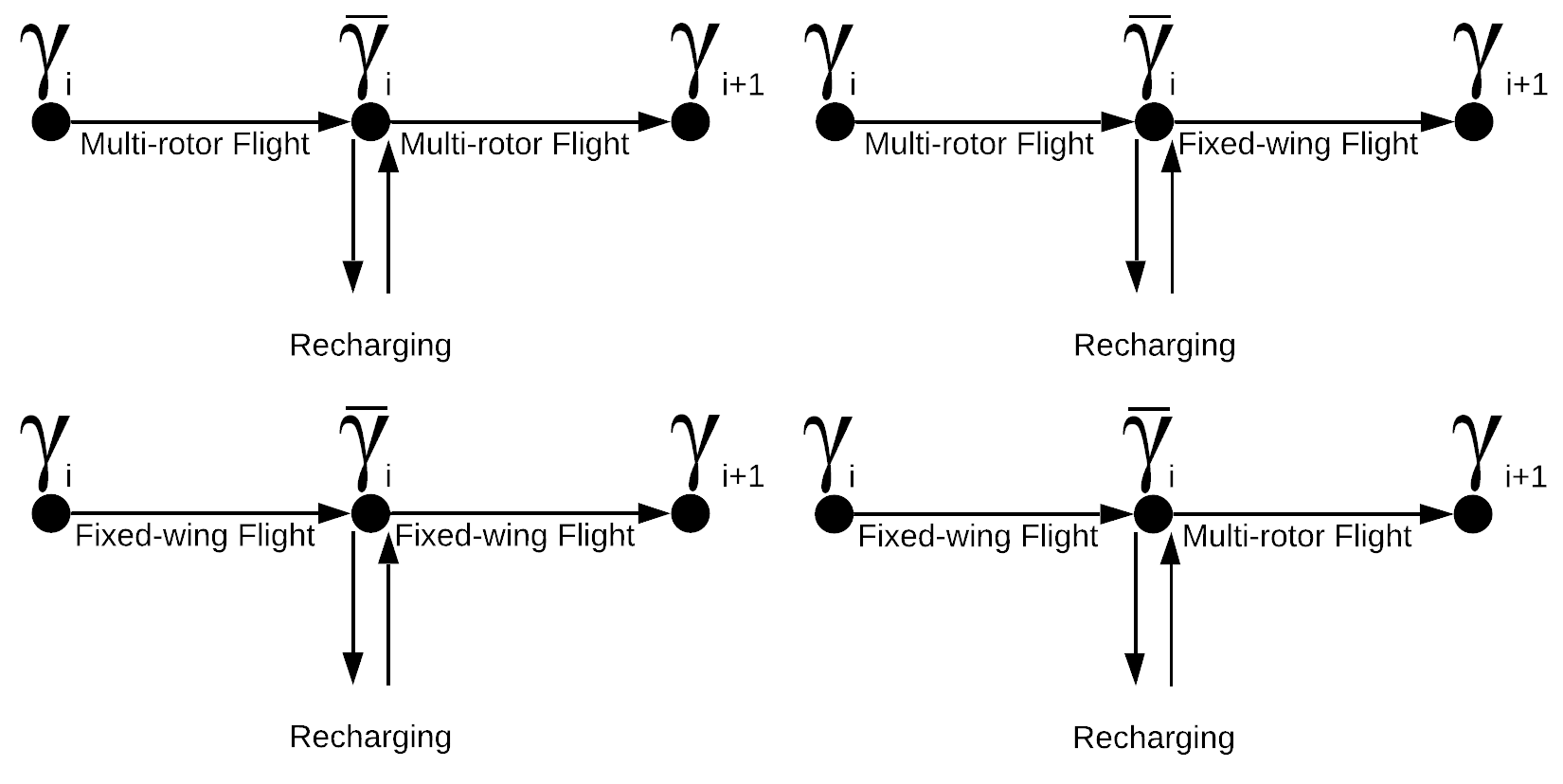}
\caption{\MDUM, \FDUF, \MDUF, \FDUM.}
\label{fig:xdux}
\end{figure}

\begin{figure}[t]
\centering
\includegraphics[width = \columnwidth]{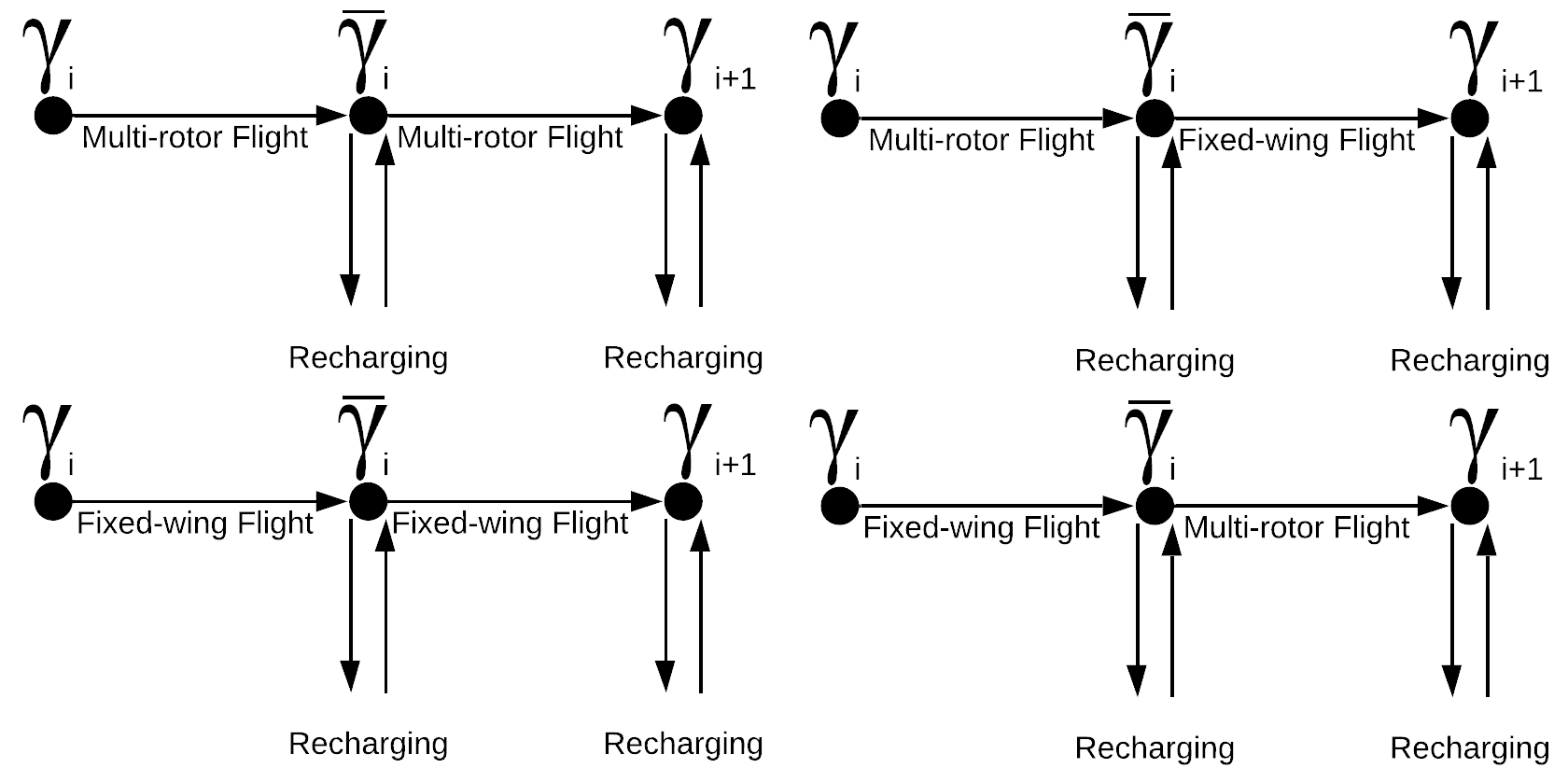}
\caption{\MDUMDU, \FDUFDU, \MDUFDU, \FDUMDU.}
\label{fig:xduxdu}
\end{figure}

Equation~\ref{eq:costPerOne} gives the cost of traveling between two entry sites of different \BCS. The actual cost depends on the five binary indicator variables: $I_1(\gamma_j,\overline{\gamma}_j)$, $I_2(\gamma_j)$, $I_3(\overline{\gamma_j})$, $I_1(\overline{\gamma}_j,\gamma_{j+1})$, and $I_2(\gamma_{j+1})$. This gives a total of $2^5$ possible travel options. However, fourteen of these thirty-two options are redundant. Specifically, if the UAV chooses to recharge while traveling on the UGV, then also recharging on either end of this leg is redundant and, in fact, more time-consuming since it will have to take-off and land more than once. Formally, if $I_3(\overline{\gamma_j})=1$, then the optimal algorithm will never set $I_2(\gamma_j)=1$ nor $I_2(\gamma_{j+1})=1$. Also if $I_3(\overline{\gamma_j})=1$ then what $I_1(\gamma_j,\overline{\gamma}_j)$ and $I_1(\overline{\gamma}_j,\gamma_{j+1})$ is equal to does not matter and we can eliminate those possibilities as well. We leave two states where $I_1(\gamma_j,\overline{\gamma}_j)=1$ and $I_1(\gamma_j,\overline{\gamma}_j)=0$ to handle the state in which the UAV will use the UGV as a transport. Therefore, of these $2^5$ possibilities, we can eliminate fourteen, leaving a total of eighteen possibilities. These are shown in Figures~\ref{fig:xx}, ~\ref{fig:xdtu}, ~\ref{fig:xxdu}, ~\ref{fig:xdux}, ~\ref{fig:xduxdu}. Note that since we assume that the UAV starts with 100\% battery capacity, we will never recharge at the first entry site. 

We denote the eighteen edge combinations using the notation: M = Multi-rotor, F = Fixed-wing, D = Down/Land, U = Up/Take-off, and T = Transit. The first leg is always the UAV flying to cover the \BC. We describe the actual edge costs in the remainder of this section.

In the following, we show how to compute the edge cost between vertices $v_{a_i}^{k_i}$ and $v_{a_j}^{k_j}$. $k'_i$ denotes the battery at $v_{b_i}^{k'_i}$ if going from $v_{a_i}^{k_i}$ and $v_{a_j}^{k_j}$. Note that there will also be edges between $v_{b_i}^{k_i}$ and $v_{a_j}^{k_j}$, $v_{a_i}^{k_i}$ and $v_{b_j}^{k_j}$, and $v_{b_i}^{k_i}$ and $v_{b_j}^{k_j}$. The costs for these edges can be obtained using the same formula just by swapping $a$ with $b$ and vice-versa.

The cost of the \MM, \FF, \MF, and \FM type of edges, Figure~\ref{fig:xx}, between $v_{a_i}^{k_i}$ and $v_{a_j}^{k_j}$ is $\infty$ if the energy required to go from $a_i$ to $b_i$ and then to $a_j$ is more than $k_j-k_i$. Else, the edge cost is given by:
\begin{equation}\label{eq:MM}
\begin{split}
&T_{\MM}(v_{a_i}^{k_i}, v_{a_j}^{k_j}) = t_{M}(v_{a_i}^{k_i},v_{b_i}^{k'_i}) + t_{M}(v_{b_i}^{k'_i},v_{a_j}^{k_j}),
\end{split}
\end{equation}
\begin{equation}\label{eq:FF}
\begin{split}
&T_{\FF}(v_{a_i}^{k_i}, v_{a_j}^{k_j}) = t_{F}(v_{a_i}^{k_i},v_{b_i}^{k'_i}) + t_{F}(v_{b_i}^{k'_i},v_{a_j}^{k_j}).
\end{split}
\end{equation}
\begin{equation}\label{eq:MF}
\begin{split}
&T_{\MF}(v_{a_i}^{k_i}, v_{a_j}^{k_j}) = t_{M}(v_{a_i}^{k_i},v_{b_i}^{k'_i}) + t_{F}(v_{b_i}^{k'_i},v_{a_j}^{k_j}),
\end{split}
\end{equation}
\begin{equation}\label{eq:FM}
\begin{split}
&T_{\FM}(v_{a_i}^{k_i}, v_{a_j}^{k_j}) = t_{F}(v_{a_i}^{k_i},v_{b_i}^{k'_i}) + t_{M}(v_{b_i}^{k'_i},v_{a_j}^{k_j}),
\end{split}
\end{equation}

The cost of the \MDTU and \FDTU type of edges, Figure~\ref{fig:xdtu}, between $v_{a_i}^{k_i}$ and $v_{a_j}^{k_j}$ is equal to $\infty$ if the energy required to go from $a_i$ to $b_i$ is more than $k'_i-k_i$. Else, the edge cost is given by:
\begin{equation}\label{eq:MDTU}
\begin{split}
&T_{\MDTU}(v_{a_i}^{k_i}, v_{a_j}^{k_j}) = t_{M}(v_{a_i}^{k_i},v_{b_i}^{k'_i}) + t_{L}\\
& + \max(t_{G}(v_{b_i}^{k'_i},v_{a_j}^{k_j}), r \cdot e) + t_{TO},
\end{split}
\end{equation}
\begin{equation}\label{eq:FDTU}
\begin{split}
&T_{\FDTU}(v_{a_i}^{k_i}, v_{a_j}^{k_j}) = t_{F}(v_{a_i}^{k_i},v_{b_i}^{k'_i}) + t_{L}\\
& + \max(t_{G}(v_{b_i}^{k'_i},v_{a_j}^{k_j}), r \cdot e) + t_{TO},
\end{split}
\end{equation}
where $e = \max\{0,k_j - (k'_i - \norm{b_i-a_j})\}$ gives the recharging amount.

The cost of the \MMDU, \FFDU, \MFDU, and \FMDU type of edges, Figure~\ref{fig:xxdu}, between $v_{a_i}^{k_i}$ and $v_{a_j}^{k_j}$ is $\infty$ if the energy required to go from $a_i$ to $b_i$ and then to $a_j$ is more than $k_j-k_i$. Else, the edge cost is:
\begin{equation}\label{eq:MMDU}
\begin{split}
&T_{\MMDU}(v_{a_i}^{k_i}, v_{a_j}^{k_j}) = t_{M}(v_{a_i}^{k_i},v_{b_i}^{k'_i}) + t_{M}(v_{b_i}^{k'_i},v_{a_j}^{k_j}) + t_{L}\\
& + r \cdot e + t_{TO},
\end{split}
\end{equation}
\begin{equation}\label{eq:FFDU}
\begin{split}
&T_{\FFDU}(v_{a_i}^{k_i}, v_{a_j}^{k_j}) = t_{F}(v_{a_i}^{k_i},v_{b_i}^{k'_i}) + t_{F}(v_{b_i}^{k'_i},v_{a_j}^{k_j}) + t_{L}\\
& + r \cdot e + t_{TO},
\end{split}
\end{equation}
\begin{equation}\label{eq:MFDU}
\begin{split}
&T_{\MFDU}(v_{a_i}^{k_i}, v_{a_j}^{k_j}) = t_{M}(v_{a_i}^{k_i},v_{b_i}^{k'_i}) + t_{F}(v_{b_i}^{k'_i},v_{a_j}^{k_j}) + t_{L}\\
& + r \cdot e + t_{TO},
\end{split}
\end{equation}
\begin{equation}\label{eq:FMDU}
\begin{split}
&T_{\FMDU}(v_{a_i}^{k_i}, v_{a_j}^{k_j}) = t_{F}(v_{a_i}^{k_i},v_{b_i}^{k'_i}) + t_{M}(v_{b_i}^{k'_i},v_{a_j}^{k_j}) + t_{L}\\
& + r \cdot e + t_{TO},
\end{split}
\end{equation}
where $e = \max\{0,k_j - (k_i - (\norm{a_i-b_i} + \norm{b_i-a_j}))\}$ gives the recharging amount.

The cost of the \MDUM, \FDUF, \MDUF, and \FDUM type of edges, Figure~\ref{fig:xdux}, between $v_{a_i}^{k_i}$ and $v_{a_j}^{k_j}$ is equal to $\infty$ if the energy required to go from $a_i$ to $b_i$ is more than $k'_i-k_i$ and then $b_i$ to $a_j$ is more than $k_j-k'_i$. Else, the edge cost is given by:
\begin{equation}\label{eq:MDUM}
\begin{split}
&T_{\MDUM}(v_{a_i}^{k_i}, v_{a_j}^{k_j}) = t_{M}(v_{a_i}^{k_i},v_{b_i}^{k'_i}) + t_{L} + r \cdot e\\
& + t_{TO} + t_{M}(v_{b_i}^{k'_i},v_{a_j}^{k_j}),
\end{split}
\end{equation}
\begin{equation}\label{eq:FDUF}
\begin{split}
&T_{\FDUF}(v_{a_i}^{k_i}, v_{a_j}^{k_j}) = t_{F}(v_{a_i}^{k_i},v_{b_i}^{k'_i}) + t_{L} + r \cdot e\\
& + t_{TO} + t_{F}(v_{b_i}^{k'_i},v_{a_j}^{k_j}),
\end{split}
\end{equation}
\begin{equation}\label{eq:MDUF}
\begin{split}
&T_{\MDUF}(v_{a_i}^{k_i}, v_{a_j}^{k_j}) = t_{M}(v_{a_i}^{k_i},v_{b_i}^{k'_i}) + t_{L} + r \cdot e\\
& + t_{TO} + t_{F}(v_{b_i}^{k'_i},v_{a_j}^{k_j}),
\end{split}
\end{equation}
\begin{equation}\label{eq:FDUM}
\begin{split}
&T_{\FDUM}(v_{a_i}^{k_i}, v_{a_j}^{k_j}) = t_{F}(v_{a_i}^{k_i},v_{b_i}^{k'_i}) + t_{L} + r \cdot e\\
& + t_{TO} + t_{M}(v_{b_i}^{k'_i},v_{a_j}^{k_j}),
\end{split}
\end{equation}
where $e = \max\{0,k'_i - (k_i - \norm{a_i-b_i})\}$ gives the recharging amount.

The cost of the \MDUMDU, \FDUFDU, \MDUFDU, and \FDUMDU type of edges, Figure~\ref{fig:xduxdu}, between $v_{a_i}^{k_i}$ and $v_{a_j}^{k_j}$ is equal to $\infty$ if the energy required to go from $a_i$ to $b_i$ is more than $k'_i-k_i$ and then $b_i$ to $a_j$ is more than $k_j-k'_i$. Else, the edge cost is given by:
\begin{equation}\label{eq:MDUMDU}
\begin{split}
&T_{\MDUMDU}(v_{a_i}^{k_i}, v_{a_j}^{k_j}) = t_{M}(v_{a_i}^{k_i},v_{b_i}^{k'_i}) + t_{L} + r \cdot e_1 + t_{TO}\\
& + t_{M}(v_{b_i}^{k'_i},v_{a_j}^{k_j}) + t_{L} + r \cdot e_2 + t_{TO},
\end{split}
\end{equation}
\begin{equation}\label{eq:FDUFDU}
\begin{split}
&T_{\FDUFDU}(v_{a_i}^{k_i}, v_{a_j}^{k_j}) = t_{F}(v_{a_i}^{k_i},v_{b_i}^{k'_i}) + t_{L} + r \cdot e_1 + t_{TO}\\
& + t_{F}(v_{b_i}^{k'_i},v_{a_j}^{k_j}) + t_{L} + r \cdot e_2 + t_{TO},
\end{split}
\end{equation}
\begin{equation}\label{eq:MDUFDU}
\begin{split}
&T_{\MDUFDU}(v_{a_i}^{k_i}, v_{a_j}^{k_j}) = t_{M}(v_{a_i}^{k_i},v_{b_i}^{k'_i}) + t_{L} + r \cdot e_1 + t_{TO}\\
& + t_{F}(v_{b_i}^{k'_i},v_{a_j}^{k_j}) + t_{L} + r \cdot e_2 + t_{TO},
\end{split}
\end{equation}
\begin{equation}\label{eq:FDUMDU}
\begin{split}
&T_{\FDUMDU}(v_{a_i}^{k_i}, v_{a_j}^{k_j}) = t_{F}(v_{a_i}^{k_i},v_{b_i}^{k'_i}) + t_{L} + r \cdot e_1 + t_{TO}\\
& + t_{M}(v_{b_i}^{k'_i},v_{a_j}^{k_j}) + t_{L} + r \cdot e_2 + t_{TO},
\end{split}
\end{equation}
where $e_1 = \max\{0,k'_i - (k_i - \norm{a_i-b_i})\}$ and $e_2 = \max\{0,k_j - (k'_i - \norm{b_i-a_j})\}$ gives the recharging amount for $e_1$ and $e_2$, respectively.

The actual edge cost between $v_{a_i}^{k_i}$ and $v_{a_j}^{k_j}$ is the minimum of all eighteen types. Specifically, the final edge cost is given by:
\begin{equation}\label{eq:totCost}
\begin{split}
&T(v_{a_i}^{k_i}, v_{a_j}^{k_j}) = \min \{ T_{\text{\MM}}, T_{\text{\FF}}, T_{\text{\MF}}, T_{\text{\FM}}, T_{\text{\MDTU}}, T_{\text{\FDTU}}, \\
&T_{\text{\MMDU}}, T_{\text{\FFDU}}, T_{\text{\MFDU}}, T_{\text{\FMDU}}, T_{\text{\MDUM}}, T_{\text{\FDUF}}, T_{\text{\MDUF}},\\
&T_{\text{\FDUM}}, T_{\text{\MDUMDU}}, T_{\text{\FDUFDU}}, T_{\text{\MDUFDU}}, T_{\text{\FDUMDU}} \}.
\end{split}
\end{equation}
We also keep track of which type of edge gives the final edge cost. This is used when converting the GTSP tour into a solution for the original problem.

\subsection{Solving GTSP}\label{sec:glns}
We solve the GTSP instance using the GLNS solver~\cite{Smith2016GLNS}. GLNS uses a neighborhood search heuristic to find the optimal solution for the given GTSP instance. GLNS also allows for finding feasible solutions in lesser time, potentially at the expense of optimality. 

Common alternatives for finding the optimal GTSP solution are Integer Programming or reducing GTSP to TSP~\cite{noon1993efficient} and then using a TSP solver such as \CON~\cite{applegate2006concorde}. In our previous work~\cite{yu2018autonomous}, we showed that GLNS finds the optimal solution for a similar class of GTSP instances in times that are at least an order of magnitude faster than the other approaches. As a result, we focus on only using GLNS for solving the GTSP instances in this paper. 

\subsection{Converting the GTSP solution to a Coverage Tour}\label{sec:revert}
The optimal tour obtained from the GTSP solver is a tour that visits exactly one vertex in each cluster. Recall that one cluster corresponds to one \BC. The optimal tour will visit only one vertex within a cluster. The chosen vertex corresponds specifies the entry site for the \BC as well as the corresponding battery level. 

For example, if the edge between $v_{a_i}^{k_i}$ and $v_{b_j}^{k_j}$ is selected, then this implies the UAV will cover the $i^{th}$ \BC with $a_i$ as the entry site and $b_i$ as the exit site. Then, the UAV will travel from $b_i$ to the entry site of the next \BC which is chosen to be $b_j$. The actual mode of transportation between $v_{a_i}^{k_i}$ and $v_{b_j}^{k_j}$ depends on the type of the edge, denoted by the eighteen edges in Figures~\ref{fig:xx}, ~\ref{fig:xdtu}, ~\ref{fig:xxdu}, ~\ref{fig:xdux}, ~\ref{fig:xduxdu}. Depending the type, we construct the actual tour and recharging schedule for the UAV. 

We can determine the UGV path based on the type of edges chosen by considering the edges in the order they appear in the optimal GTSP tour. For a \MM, \MF, \FM, \FF edge, the UGV is not required. For an \MDUM, \FDUF, \MDUF, \FDUM edge, we add the exit site of the first \BC to the UGV tour. Similarly, for an \MMDU, \FFDU, \MFDU, \FMDU, we add the entry site of the second \BC to the UGV tour. Finally, for \MDUMDU, \FDUFDU, \MDUFDU, \FDUMDU, \MDTU, and \FDTU edges, we add the exit site of the first \BC and the entry site of the second one to the UGV tour.

\subsection{Performance Guarantees}
If the GTSP solver finds the optimal tour, then the corresponding UAV tour is also the optimal solution to Problem~\ref{pr:MPC} with the additional assumption that the UGV is as fast as the UAV. If the UAV is faster than the UGV, then it is possible that the solution yields paths where the UAV reaches a landing site before the UGV. In such cases, one possibility is to have more than one UGV that can support the UAV tour. In our previous work~\cite{yu2019algorithms}, we presented an Integer Programming solution that minimizes the number of slower UGVs required to support the UAV tour. 

\section{Simulations}\label{sec:sim}
In this section, we present qualitative and quantitative results to evaluate the proposed algorithm. In particular, we analyze the effect of various parameters on the tour cost and the computational time of the algorithm. The experiments were run on an Ubuntu 16.04 computer with an Intel i7-8750H CPU running at 2.2GHz, with 6 physical cores, 32GB of RAM, and a GTX 1070 GPU.

\subsection{Qualitative Examples}
\begin{figure}
    \centering
    \includegraphics[width = 0.8\columnwidth]{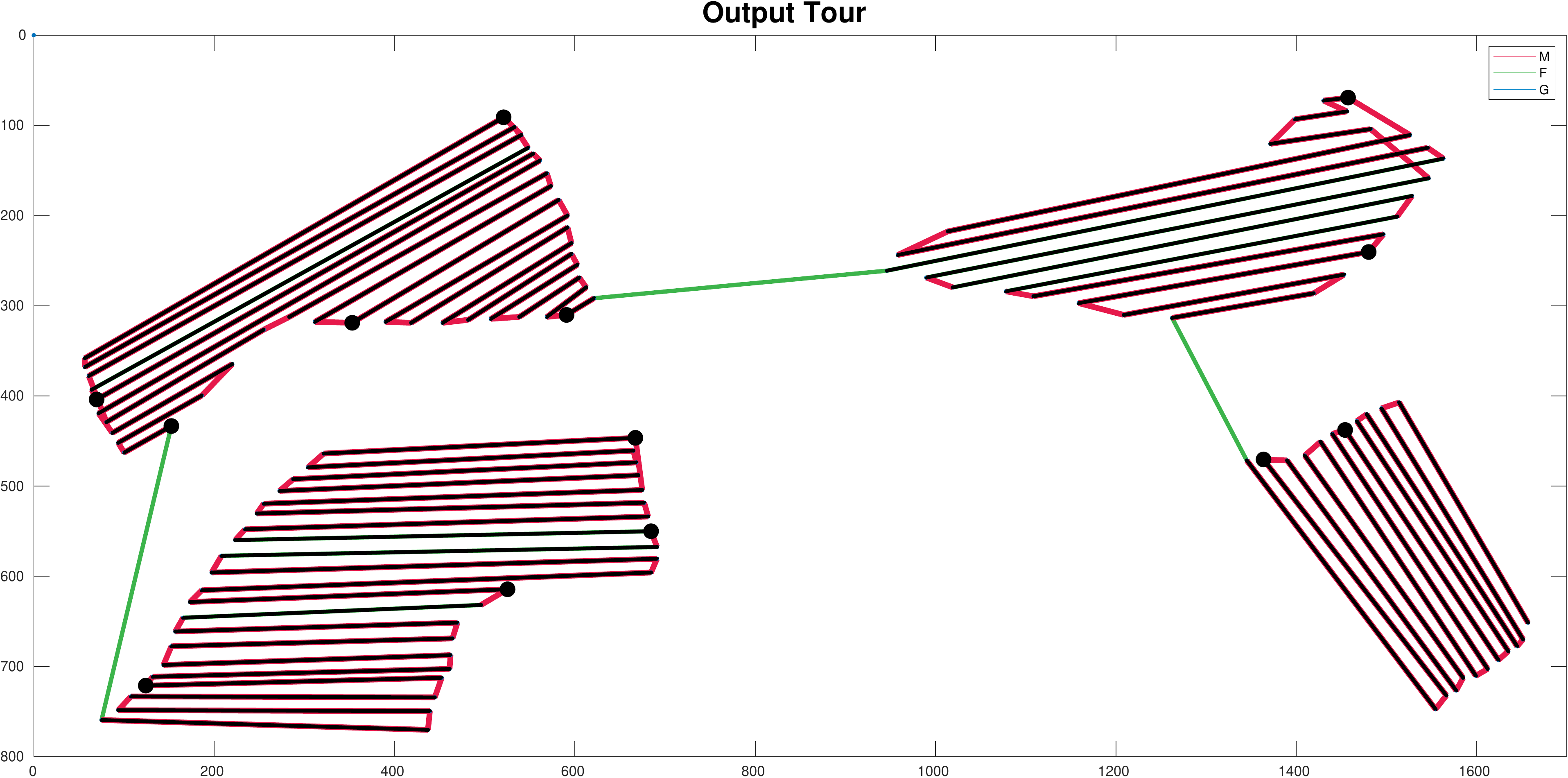}
    \caption{Solution to the input instance given in Figure~\ref{fig:BCS}.}
    \label{fig:BCSOutput}
\end{figure}

We use the \BCS given in Figure~\ref{fig:BCS} as the input. There are a total of 66 \BCS. The solution is shown in Figure~\ref{fig:BCSOutput}. The \BCS are marked with rectangles. The input parameters were set to: $t_{TO} = 5$, $t_{L} = 45$, $r = 2$, $D_{\max} = 1800$, $C = 20$, $fRatio = 3$, and $TR = 3$. The UGV speed was set to be 20\% as that of the UAV.
 
\begin{figure*}[htb]
\centering{
\subfigure[Input]{\includegraphics[width=0.28\textwidth]{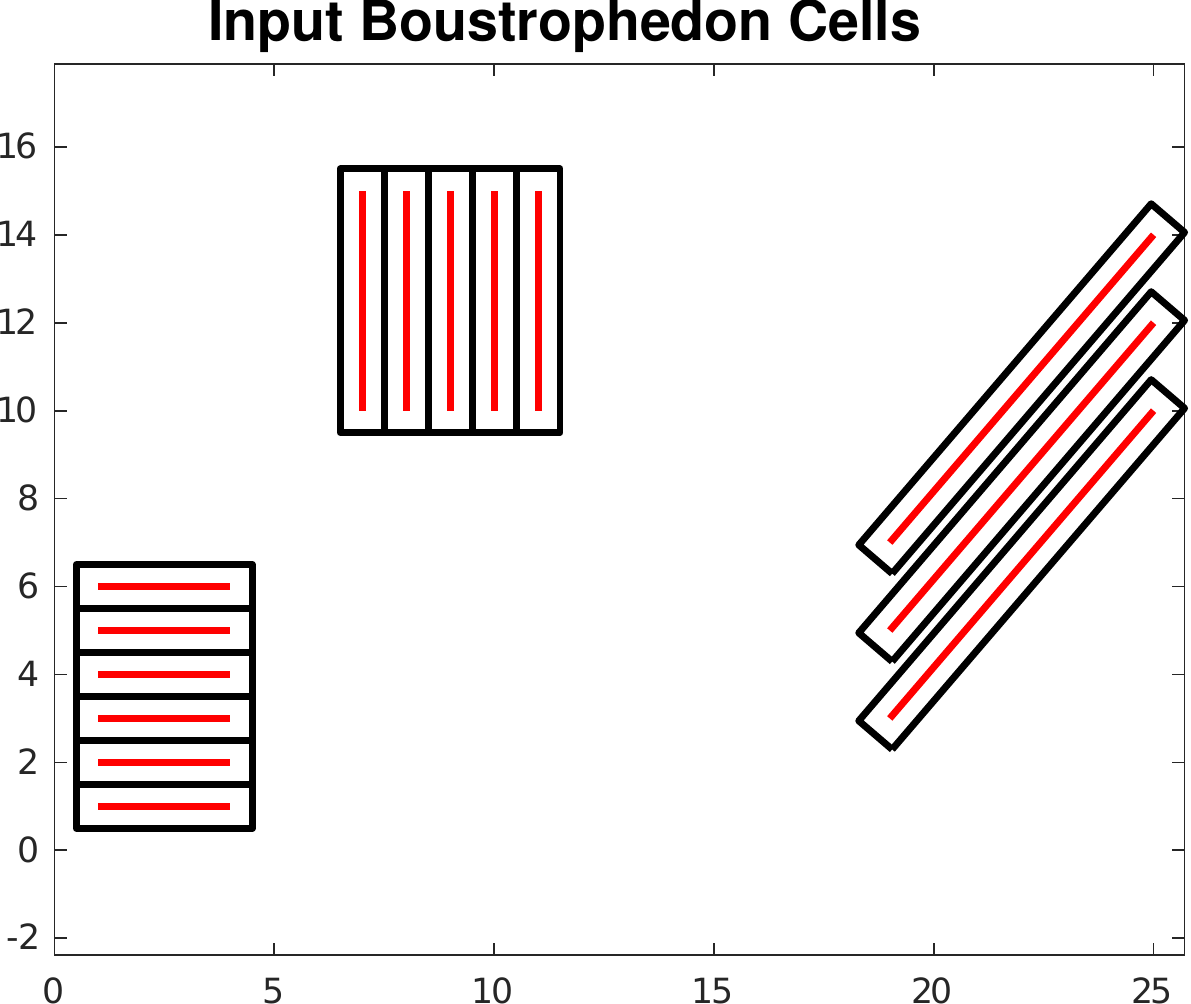}\label{fig:Input}}
\subfigure[Multi-rotor only with input $t_{TO} = 5$, $t_{L} = 45$, $r = 2$, $D_{\max} = 200$, UGV speed is one-fifth of the UAV, $C = 20$, $fRatio = 1$, and $TR = 5$]{\includegraphics[width=0.32\textwidth]{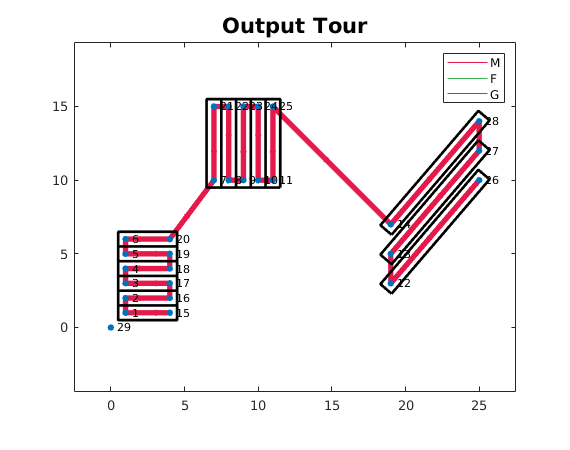}\label{fig:M}}
\subfigure[Fixed-wing only with input $t_{TO} = 5$, $t_{L} = 45$, $r = 2$, $D_{\max} = 10$, UGV speed is one-fifth of the UAV, $C = 20$, $fRatio = 50$, and $TR = 1$]{\includegraphics[width=0.32\textwidth]{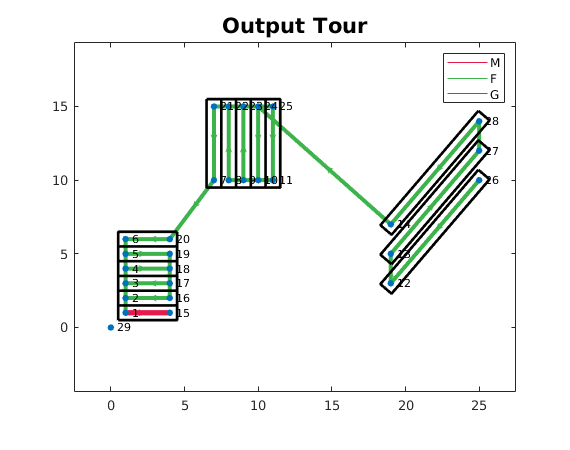}\label{fig:F}}
\subfigure[Multi-rotor with stationary recharging only with input $t_{TO} = 5$, $t_{L} = 45$, $r = 2$, $D_{\max} = 50$, UGV speed is one-fiftieth of the UAV, $C = 20$, $fRatio = 1$, and $TR = 5$]{\includegraphics[width=0.32\textwidth]{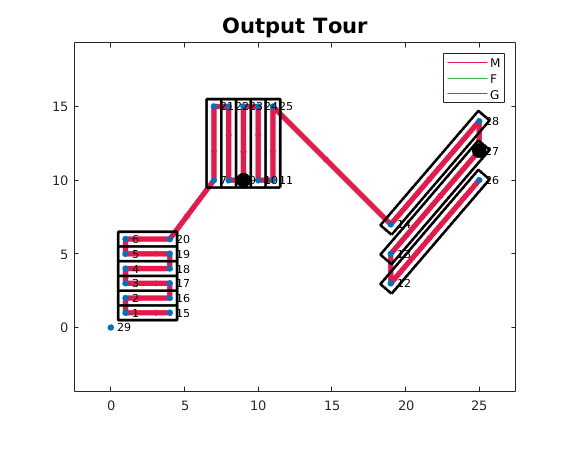}\label{fig:MLTO}}
\subfigure[Fixed-wing with stationary recharging only with input $t_{TO} = 5$, $t_{L} = 45$, $r = 2$, $D_{\max} = 20$, UGV speed is one-hundredth of the UAV, $C = 50$, $fRatio = 6$, and $TR = 0.5$ ]{\includegraphics[width=0.32\textwidth]{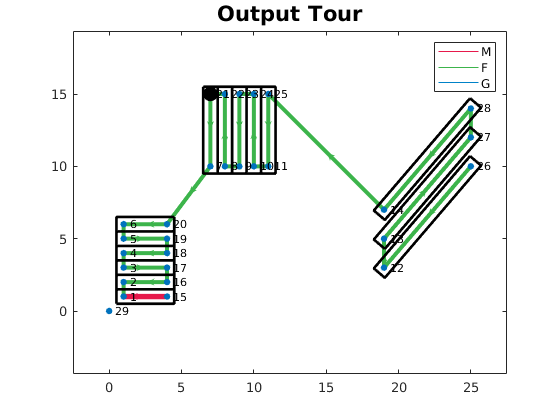}\label{fig:FLTO}}
\subfigure[Mixed flight with stationary recharging only with input $t_{TO} = 5$, $t_{L} = 45$, $r = 2$, $D_{\max} = 20$, UGV speed is one-fiftieth of the UAV, $C = 20$, $fRatio = 3$, and $TR = 1$]{\includegraphics[width=0.32\textwidth]{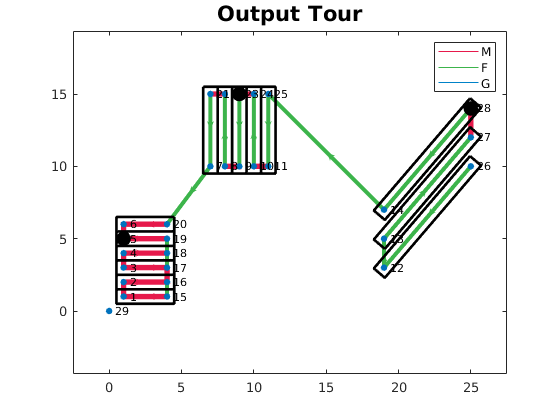}\label{fig:ALTO}}
\subfigure[UGV recharging only with input $t_{TO} = 5$, $t_{L} = 45$, $r = 2$, $D_{\max} = 50$, UGV speed is one-fifth of the UAV, $C = 20$, $fRatio = 1$, and $TR = 5$]{\includegraphics[width=0.32\textwidth]{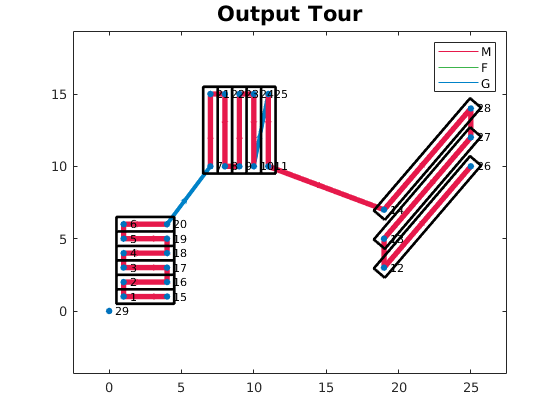}\label{fig:UGV}}
\caption{\ref{fig:Input} Input for qualitative examples to help explain input parameters and the effects.
\ref{fig:M} results in a tour that uses only the UAV in the multi-rotor configuration.
\ref{fig:F} results in a tour that uses only the UAV in the fixed-wing configuration.
\ref{fig:MLTO} Minimum number of landings/take-offs in place for the given input parameters while staying in the multi-rotor configuration.
\ref{fig:FLTO} Minimum number of landings/take-offs in place for the given input parameters while staying in the fixed-wing configuration.
\ref{fig:ALTO} Minimum number of landings/take-offs in place for the given input parameters.
\ref{fig:UGV} Minimum number of landings/take-offs using the UGV to recharge for the given input parameters.
Note that the flight along the last \BC can be either multi-rotor or fixed-wing because there is no cost for switching between flight modes and the UAV will not have to charge at the last location.}
\label{fig:qualExamples}
}
\end{figure*}

We compare the results of our algorithm with a naive baseline. The baseline approach visits each \BC in the same order in which they appear along the boundary of the polygon. The UAV lands to recharge on the UGV only when covering the next \BC would deplete it of the remaining energy. Once on the UGV, the UAV recharges to maximum capacity. The baseline approach produces a tour which requires 29564 seconds for completion (as given by Equation~\ref{eq:costFun}) where as the proposed algorithm produces a tour which requires 25595 seconds.

Figures~\ref{fig:M}--~\ref{fig:UGV} presents additional qualitative examples that show the effect of changing multiple input parameters on the solution for the input given in Figure~\ref{fig:Input}. We observe that if the UAV has enough energy capacity $D_{\max}$ then the final tour does not use the UGV (Figure~\ref{fig:M} and Figure~\ref{fig:F}). If the UGV is significantly slower than the UAV and $D_{\max}$ is small, then the UAV only recharges in-place (Figure~\ref{fig:MLTO}, Figure~\ref{fig:FLTO}, and Figure~\ref{fig:ALTO}) and does not use \MDTU or \FDTU edges. However, if the UGV is not as slow, then the tour will use \MDTU and \FDTU edges (Figure~\ref{fig:UGV}). We present quantitative evaluation of these parameters next.


\subsection{Effect of Changing $D_{\max}$ on the Tour Cost}
\begin{figure}
\centering{
\subfigure[Input]{\includegraphics[height=0.35\textwidth]{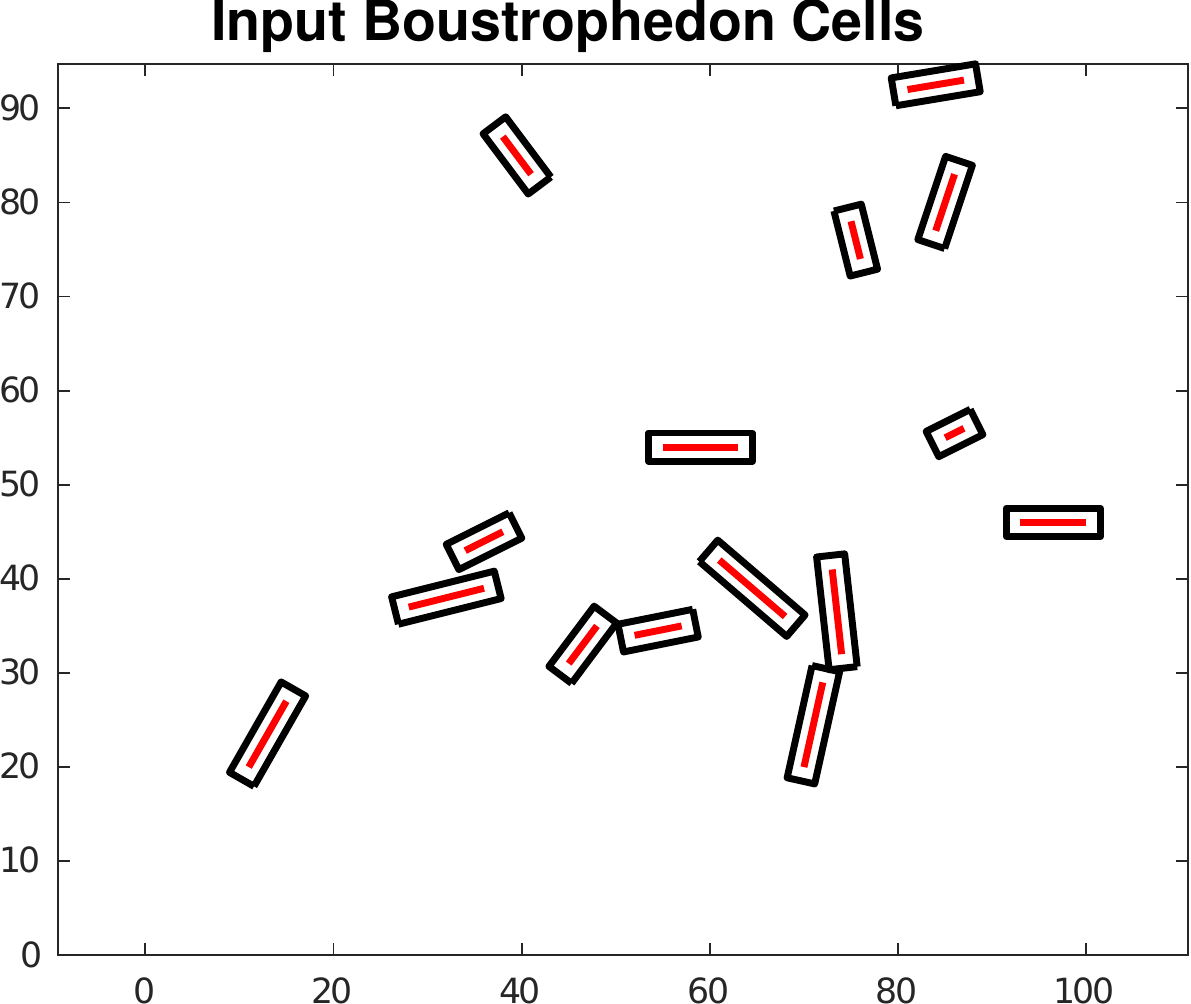}\label{fig:costBudgetInput}}
\subfigure[Cost vs. budget]{\includegraphics[height=0.35\textwidth]{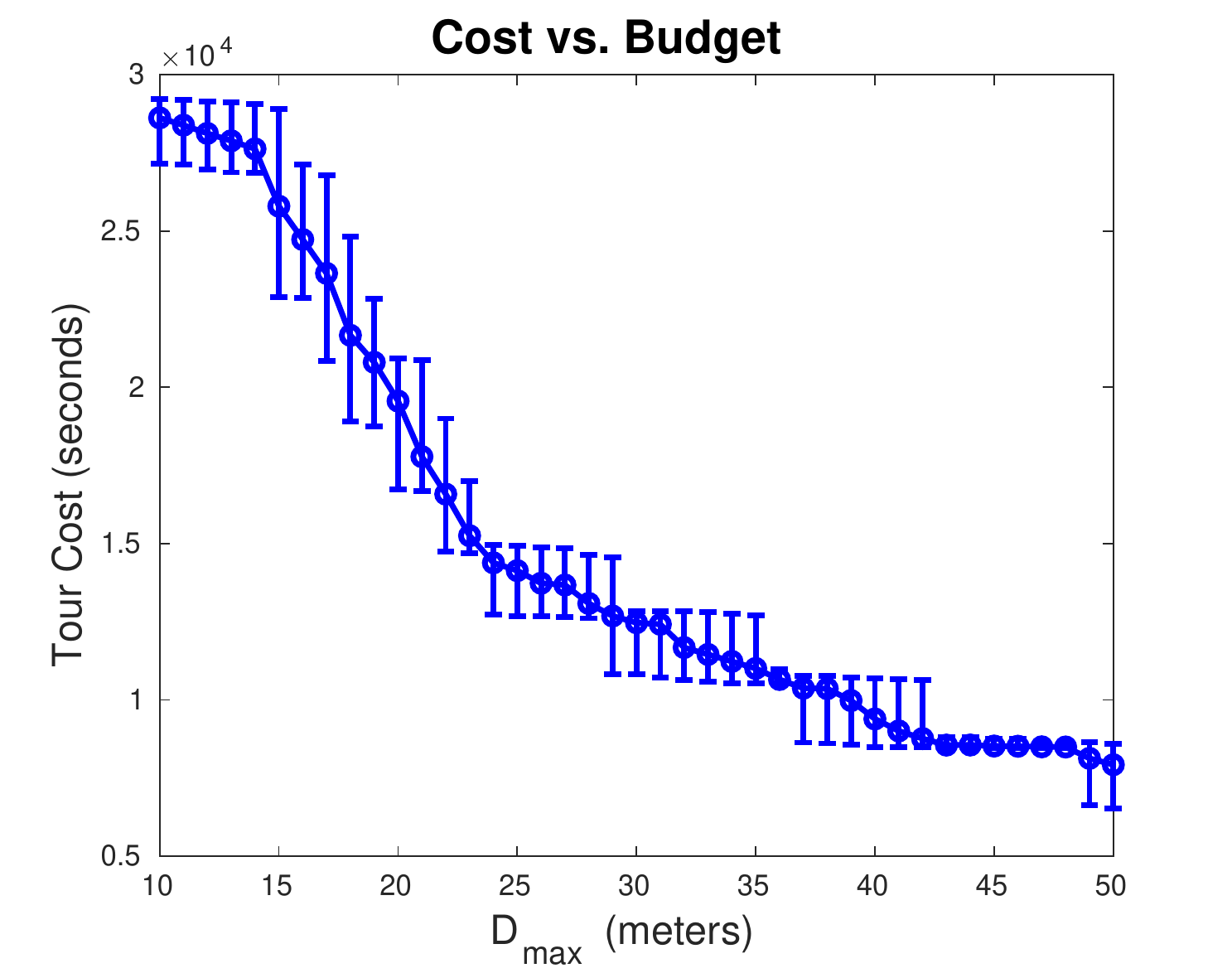}\label{fig:costBudget}}
\caption{~\ref{fig:costBudgetInput} Example input \BCS for the 10 trials used for generating~\ref{fig:costBudget}. We randomly create 15 non-overlapping \BCS, each no more than 10m.~\ref{fig:costBudget} Tour cost vs. flight budget, $D_{\max}$. We vary the total budget as well as the distance per battery level. The input parameters were: $t_{TO} = 1000$, $t_{L} = 1000$, $r = 2$, $C = 20$, $fRatio = 3$, and $TR = 3$. The UGV is 5 times slower than the UAV.}
}
\end{figure}

Next, we study the effect of changing the total battery capacity, i.e., changing $D_{\max}$, on the total tour time. We randomly generate 15 \BCS in a 100m~$\times$~100m environment such that no two \BCS intersect with each other and each \BC is no more than 10 meters long. Figure~\ref{fig:costBudgetInput} shows one example.

We vary $D_{\max}$ from 10 meters to 50 meters. We use the same set of 15 \BCS for each value of $D_{\max}$. Figure~\ref{fig:costBudget} shows the average, minimum, and maximum value of the optimal tour cost. We observe that the tour cost decreases as $D_{\max}$ increases, as is expected. We also observe a step decrease in the minimum and maximum tour costs as $D_{\max}$ increases. This can be attributed to the fact that as $D_{\max}$ increases the UAV can travel a larger distance without running out of energy. Therefore, it may need to land/take-off fewer number of times. Each landing and taking-off operation costs a fixed amount of time. Therefore, we see a step decrease in the tour cost as $D_{\max}$ increases.

\subsection{Effect on the Computational Time}
Next, we empirically analyze the computational time as a function of some of the input parameters. 

\begin{figure}[htb]
\centering{
\subfigure[]{\includegraphics[height=0.35\textwidth]{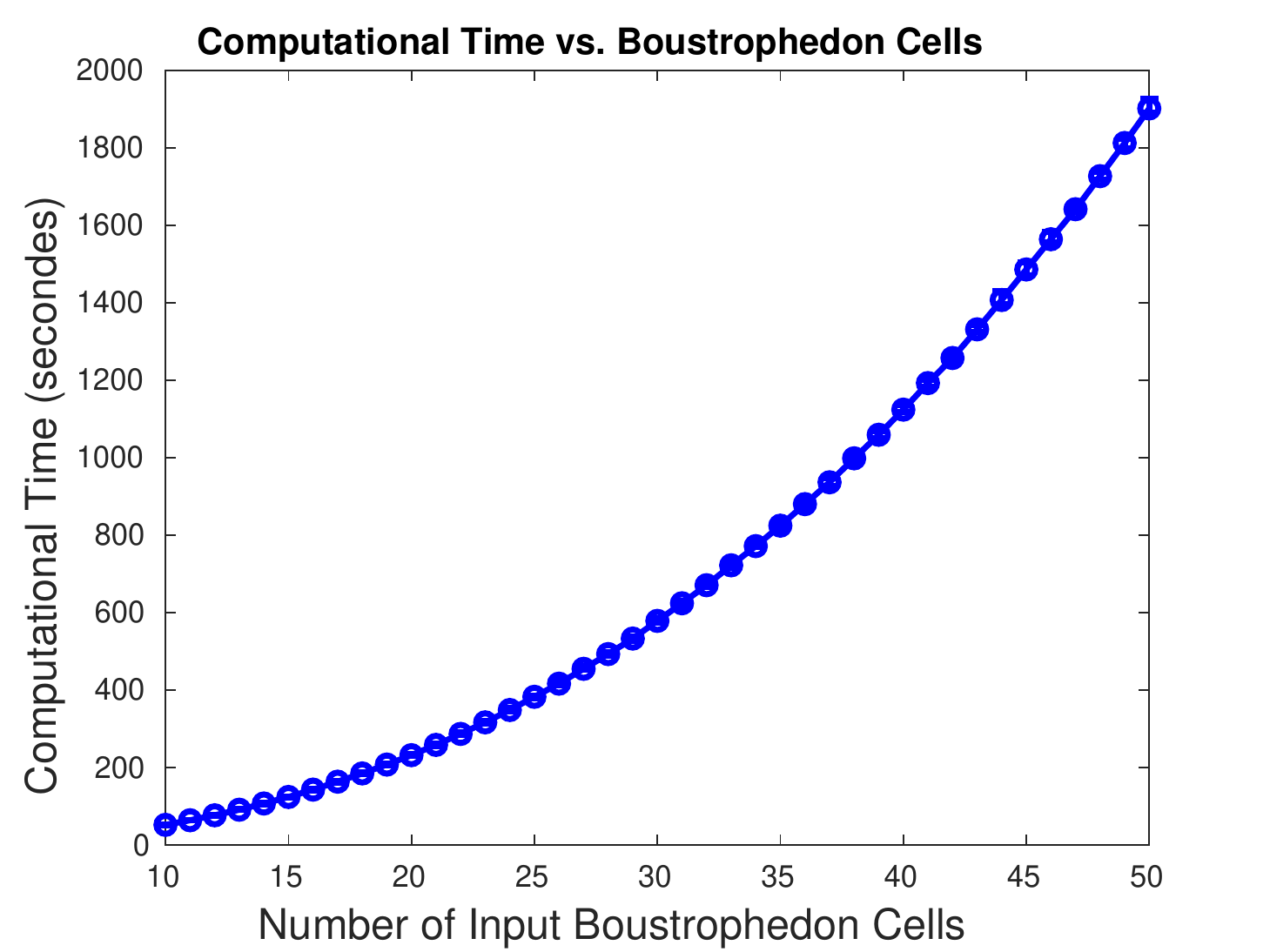}\label{fig:timeI}}
\subfigure[]{\includegraphics[height=0.35\textwidth]{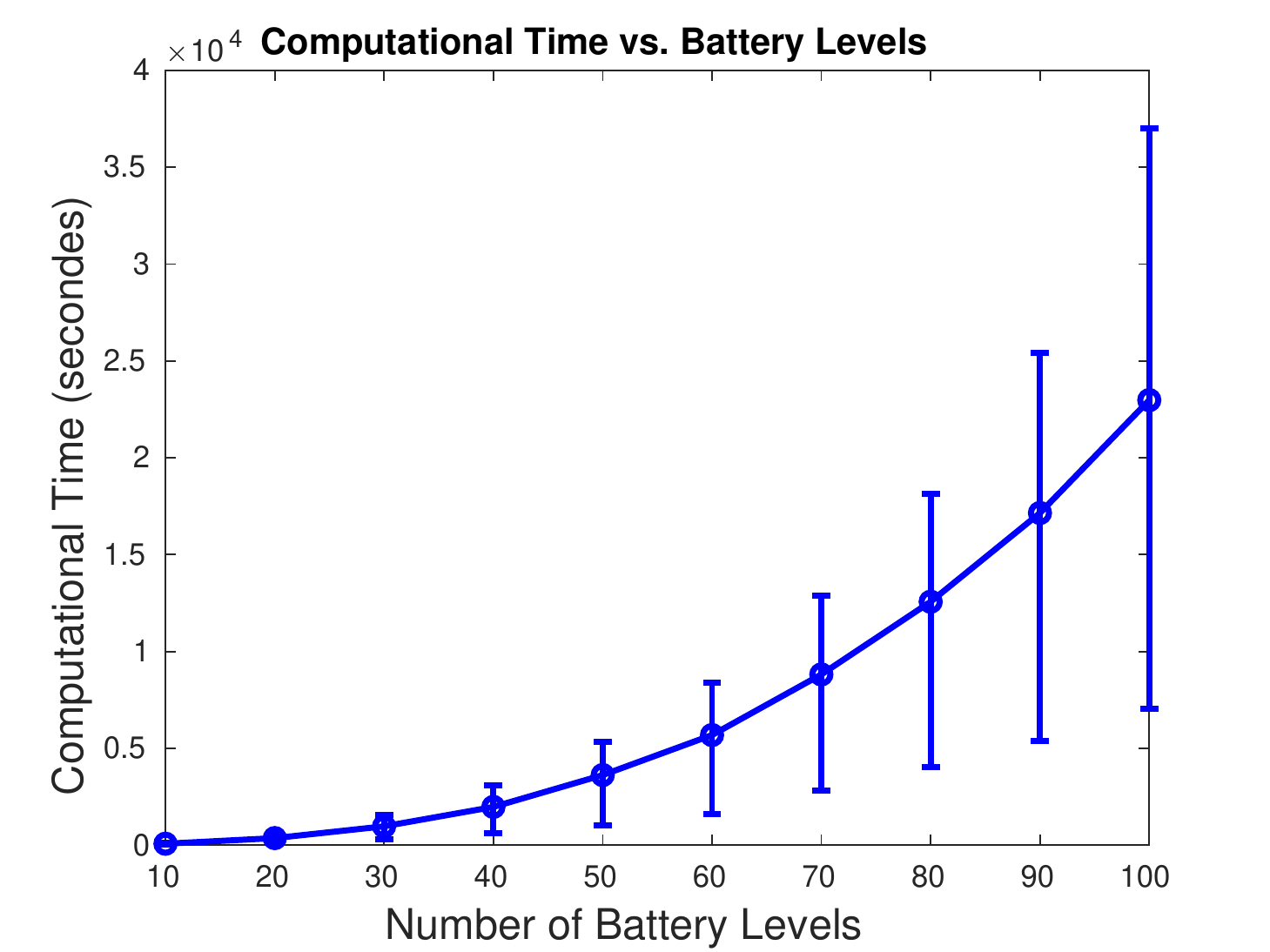}\label{fig:timeJ}}
\caption{
Input parameters: $t_{TO} = 100$, $t_{L} = 100$, $r = 2$, UGV speed is one-fifth that of the UAV, $fRatio = 3$, $TR = 1$ for both plots. 10 random sets of input \BCS were randomly generated in a $100m \times 100m$ environment. We set $C=20$ for~\ref{fig:timeI} and vary $C$ for~\ref{fig:timeJ}.}}
\end{figure}

Figure~\ref{fig:timeI} shows the effect of increasing the number of input \BCS on the computational time. The input number of \BCS is varied from 10 to 50 in steps of 1. The figure shows the average, minimum, and maximum computational times for 10 random instances. The input parameters for these experiments were kept the same: $t_{TO} = 100$, $t_{L} = 100$, $r = 2$, $C = 20$, $fRatio = 3$, and $TR = 1$.

Figure~\ref{fig:timeJ} shows the effect of increasing the battery level, i.e., $C$, on the computational time. We vary $C$ from 10 to 100 in steps of 10. The input parameters for these experiments were: $t_{TO} = 100$, $t_{L} = 100$, $r = 2$, $fRatio = 3$, and $TR = 1$. The figure shows the average, minimum, and maximum computational time for 10 random instances with 15 \BCS each.

We observe that the computational time increases (perhaps, exponentially) with increasing the number of input \BCS and battery level. Nevertheless, the computational time is still small enough (less than 50 minutes) for moderately sized instances (50 \BCS).

\section{Field Experiments}\label{sec:field}
\begin{figure}[htb]
\centering{
\subfigure[Input]{\includegraphics[height=0.24\textwidth]{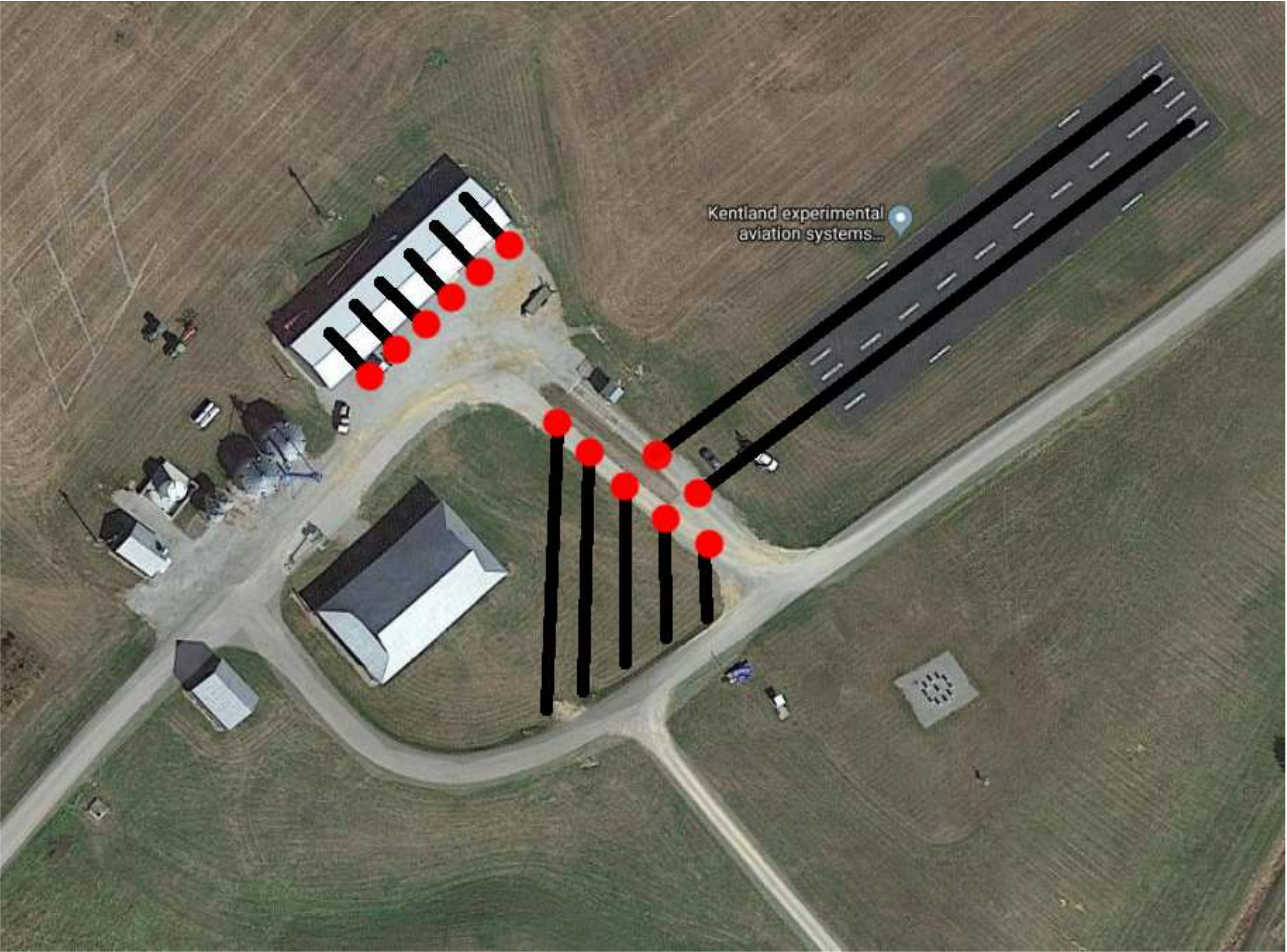}\label{fig:fieldKentInput}}
\subfigure[Output]{\includegraphics[height=0.19\textwidth]{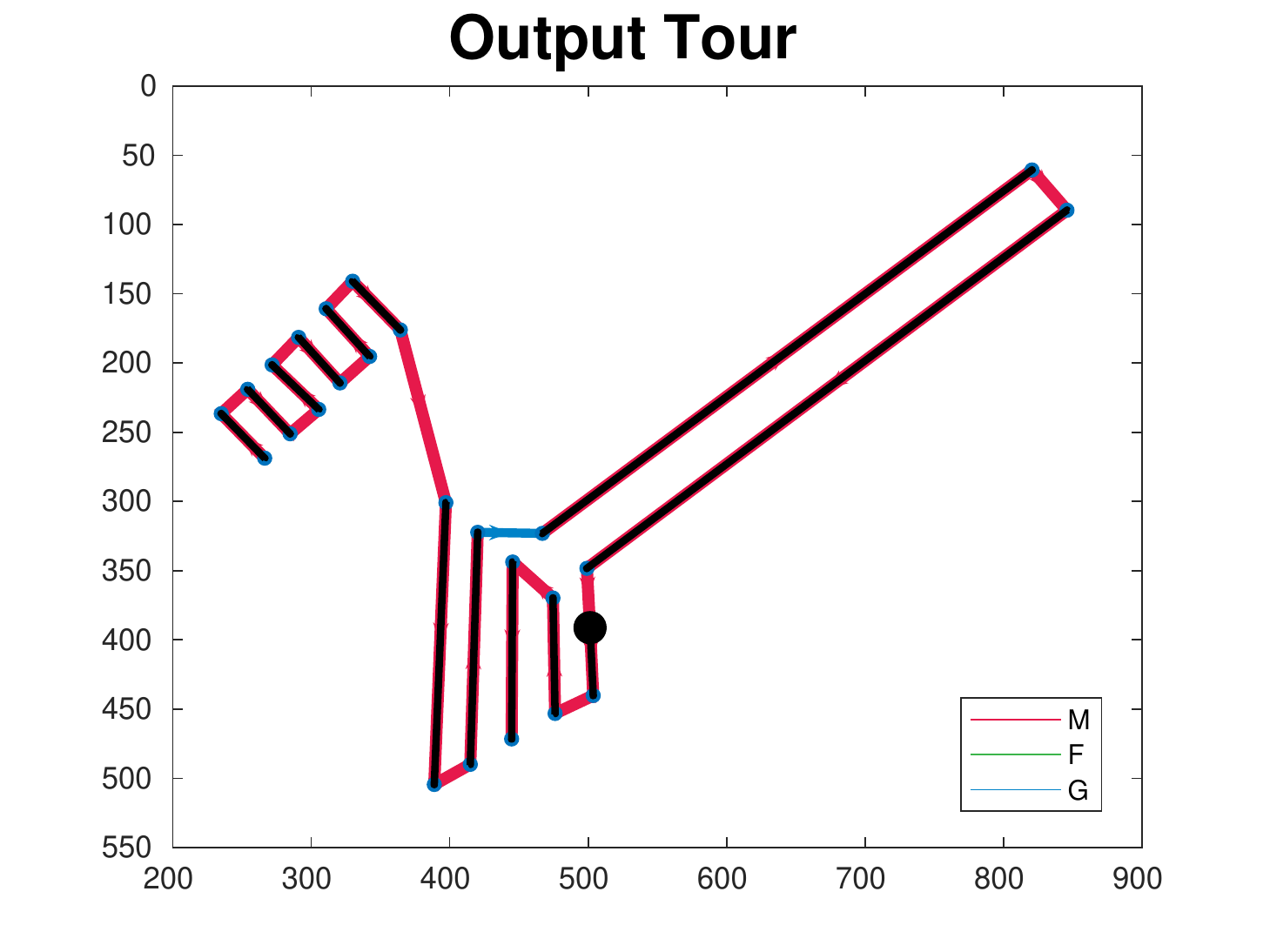}\label{fig:fieldKentOutput}}
\subfigure[Robots]{\includegraphics[height=0.19\textwidth]{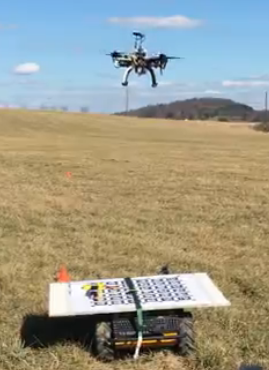}\label{fig:UAVUGV}}
\subfigure[UAV Path]{\includegraphics[height=0.17\textwidth]{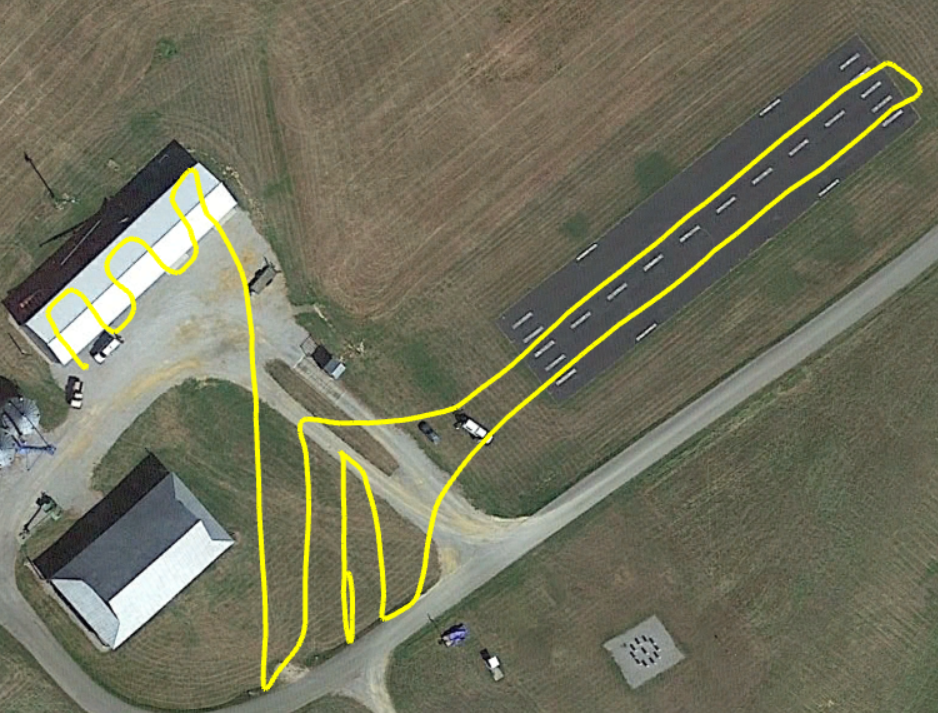}\label{fig:kentUAV}}
\subfigure[UGV Path]{\includegraphics[height=0.17\textwidth]{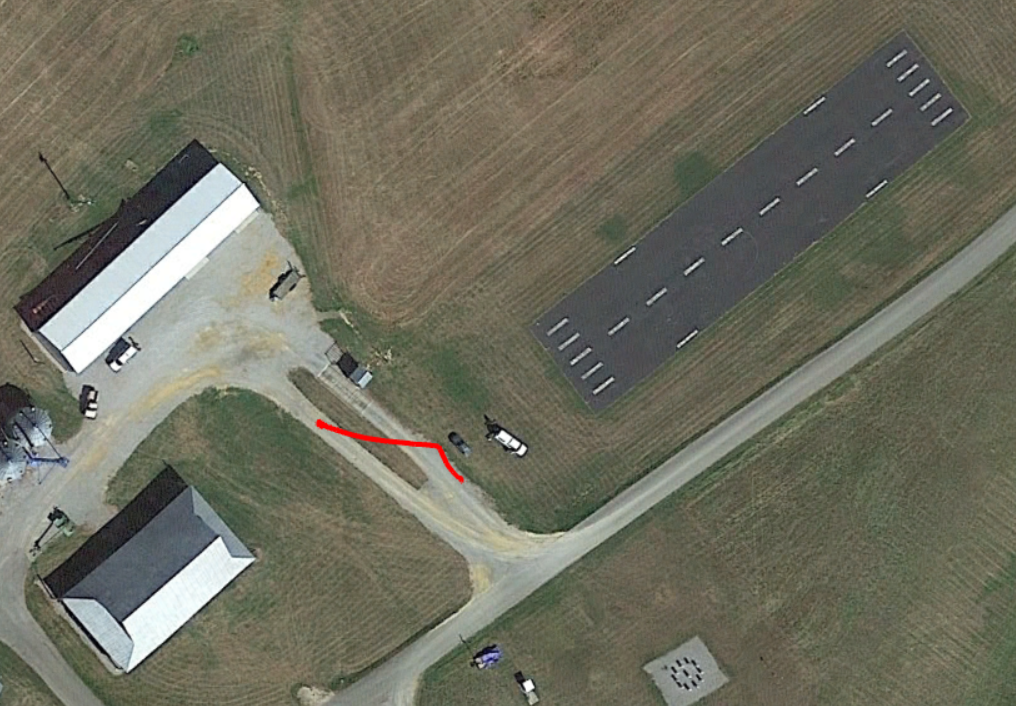}\label{fig:kentUGV}}
\caption{Proof-of-concept Experiment with 13 \BCS. The input parameters were: $D_{\max} = 1000$, $C = 100$, $t_{TO} = 100$, $t_{L} = 100$, $r = 2$, UGV speed is one-fifth that of the UAV, $fRatio = 0.5$, and $TR = 3$. The UGV is also restricted to the road network (red sites).}
}
\end{figure}

We conducted proof-of-concept field experiments using the UAV and UGV shown in Figure~\ref{fig:UAVUGV}. The UAV is a DJI 450 frame~\cite{Amazonco16:online} with a Pixhawk 2.1~\cite{Pixhawk293:online} flight controller running the APM firmware~\cite{ArduPilo3:online} and the UGV is a Clearpath Husky~\cite{HuskyUGV52:online}. The UAV is equipped with dual GPSs, a downwards facing LIDAR (for relative altitude estimation) and the IR-Lock infrared camera~\cite{IRLOCKIn21:online}. The UGV is fitted with infrared LED beacons. The IR-Lock system~\cite{IRLOCKIn21:online} allows for precision landing on the UGV with up to 10cm accuracy in nominal wind conditions. More details on the system are reported in our prior work~\cite{yu2019algorithms}.

Figure~\ref{fig:fieldKentInput} shows the input \BCS for the proof-of-concept experiment conducted at Kentland Farms at Virginia Tech. The motion of the UGV is restricted to only those sites that lie on the road. Specifically, we allow edges that conduct recharging to only occur if they correspond with sites are on the road. These sites are marked in red in Figure~\ref{fig:fieldKentInput}. The output tour for the UAV is shown in Figure~\ref{fig:fieldKentOutput}. The following parameters were used as input to the outdoor field experiments: $t_{TO} = 100$, $t_L = 100$, $r = 2$, $D_{\max} = 1000$, 13 \BCS, $C = 100$, $fRatio = 0.5$, and $TR = 3$. Since our platform is a multi-rotor, we set $fRatio < 1$. This forces the fixed-wing flights to be more expensive than the multi-rotor ones. With the addition of $TR$ our algorithm will never use any edges that use the fixed-wing mode. 

The GPS trace of the UAV and the UGV are shown in Figures~\ref{fig:kentUAV} and \ref{fig:kentUGV}. Both robots were fully autonomous during the trial that lasted 12 minutes, including taking-off and landing from the ground robot. A video of the trial is submitted as part of the multimedia attachment. The trial provides a proof-of-concept demonstration of the algorithm.

\section{Conclusion}\label{sec:con}
We present an algorithm for optimal coverage of \BCS with an energy-limited UAV and a UGV. The UGV acts as a mobile recharging station that can mule the UAV between sites, while the UAV can switch between multi-rotor and fixed-wing flight modes. We analyze the effects of various input parameters on the total tour cost as well as the computational time. We evaluate our algorithm through field experiments. 

If the UGV is slower, it is possible that the UAV may reach a site before the UGV. In~\cite{yu2019algorithms}, we showed how find the minimum number of UGVs required to ensure that the UAV can execute its tour without having to wait for the UGV. A possible extension is to find a tour for a fixed number of slower UGVs that still ensures that the UAV does not need to wait for the UGV.
We restrict the UAV to land and take-off only from the entry/exit sites of a \BC. A possible extension would be to relax this assumption which can result in even shorter tours.

\bibliographystyle{IEEEtran}
\bibliography{ref.bib}
\end{document}